\documentclass[utf8]{FrontiersinHarvard} 

\usepackage{url,hyperref,microtype}
\usepackage{multirow}
\usepackage{anyfontsize}

\def\keyFont{\fontsize{8}{11}\helveticabold }
\def\firstAuthorLast{Feng {et~al.}} 
\def\Authors{Jiale Feng\,$^{1,\dagger}$, Samuel W. Blair\,$^{2,\dagger}$, Timilehin Ayanlade\,$^{3}$, Aditya Balu\,$^{3}$, Baskar Ganapathysubramanian\,$^{3}$, Arti Singh\,$^{2}$, Soumik Sarkar\,$^{1,3,*}$, Asheesh K Singh\,$^{2,*}$}
\def\Address{$^{1}$Department of Computer Science, Iowa State University, Ames, Iowa, USA \\
$^{2}$Department of Agronomy, Iowa State University, Ames, Iowa, USA \\
$^{3}$Department of Mechanical Engineering, Iowa State University, Ames, Iowa, USA \\
$^{\dagger}$JF and SWB are co-first authors and contributed equally to this work.}
\def\corrAuthor{Corresponding Authors:}

\def\corrEmail{soumiks@iastate.edu and singhak@iastate.edu}

\begin{document}
\onecolumn
\firstpage{1}

\title[Soybean yield estimation using ground robots and DL]
{Robust soybean seed yield estimation using high-throughput ground robot videos} 

\author[\firstAuthorLast ]{\Authors} 
\address{} 
\correspondance{} 

\extraAuth{}

\maketitle

\begin{abstract}

\section{}
We present a novel method for soybean (\textit{Glycine max} (L.) Merr.) yield estimation leveraging high throughput seed counting via computer vision and deep learning techniques. Traditional methods for collecting yield data are labor-intensive, costly, prone to equipment failures at critical data collection times, and require transportation of equipment across field sites. Computer vision, the field of teaching computers to interpret visual data, allows us to extract detailed yield information directly from images. By treating it as a computer vision task, we report a more efficient alternative, employing a ground robot equipped with fisheye cameras to capture comprehensive videos of soybean plots from which images are extracted in a variety of development programs. These images are processed through the P2PNet-Yield model, a deep learning framework where we combined a Feature Extraction Module (the backbone of the P2PNet-Soy) and a Yield Regression Module to estimate seed yields of soybean plots. Our results are built on three years of yield testing plot data - 8500 in 2021, 2275 in 2022, and 650 in 2023. With these datasets, our approach incorporates several innovations to further improve the accuracy and generalizability of the seed counting and yield estimation architecture, such as the fisheye image correction and data augmentation with random sensor effects. The P2PNet-Yield model achieved a genotype ranking accuracy score of up to 83\%. It demonstrates up to a 32\% reduction in time to collect yield data as well as costs associated with traditional yield estimation, offering a scalable solution for breeding programs and agricultural productivity enhancement.

\tiny
 \keyFont{ \section{Keywords:} 
 Yield estimation, Soybean seed counting, Plant phenotyping, Deep learning, Computer vision} 
\end{abstract}

\section{Introduction}
\label{sec:intro}

Soybean (\textit{Glycine max} (L.) Merr.) is one of the most important crops in the world. It is a legume that serves as an excellent source of high protein and oil for both humans and livestock~\citep{medic2014current}. For soybean cultivar development by breeders, seed yield is one of the most critical traits for making selections and cultivar release decisions. Current methods for gathering yield data on experimental lines and candidate varieties require expensive machinery, extensive travel, and prolonged equipment operation. These are all prone to equipment breakdowns and incur high maintenance costs~\citep{singh2021plant}. The data collection procedure involves harvesting thousands, and potentially hundreds of thousands, of plots across multiple locations. These economic and time burdens have motivated researchers to explore new techniques, such as remote sensing, ground robot systems, as well as machine learning (ML) and computer vision (CV) methods, to estimate yield in a more efficient and cost-effective manner. 

Significant improvements in machine learning and computer vision have given breeders new approaches for cultivar development using remote sensing platforms and ground robot systems ~\citep{singh2021plant_ch28, sarkar2024cyber}. On the one hand, remote sensing platforms such as uncrewed aerial systems (UAS) offer data collection and phenotyping tools that can be used to estimate yield~\citep{herr2023unoccupied}. On the other hand, ground robot systems utilize ground data like field images or LiDAR data for yield prediction. This paper focuses on using ground data to estimate yield. For the ground robot systems, the detection and quantification of plant organs can serve as a proxy for crop yield. Recent developments in plant organ detection have been applied successfully in various crops, such as apple orchards for fruit detection~\citep{bargoti2017image, kang2020fast}, grape vineyards for grape and shoot detection~\citep{grimm2019adaptable, guadagna2023using}, sweet pepper fruits~\citep{sa2016deepfruits} and peanuts~\citep{puhl2021infield}. 

Applying similar detection techniques to soybean pods offers noteworthy value to soybean breeders for soybean yield estimation. Soybean pod count has been shown to strongly correlate with yield~\citep{mcguire2021}. However, unlike other crops, soybean pods present unique challenges due to their smaller size and the occlusion caused by dense foliage. As sensor technology and ML tools advance, methods for estimating soybean pod counts as yield are continually emerging. There have been several studies using deep-learning models and computer vision approaches to detect and quantify soybean pods. Some used RGB imagery with a black or white background to image pods from mature soybean plants~\citep{xiang2023yolo, zhang2023high, yu2024accurate}, while others used similar background techniques but with potted plants~\citep{lu2022soybean,he2023recognition}. Notably, ~\cite{Riera2021} proposed a deep multiview image fusion architecture that minimizes human intervention and enhances yield estimation accuracy. This method uses a deep learning (DL) framework to detect soybean pods and estimate yield from RGB images collected by a mobile ground phenotyping unit. It improves the efficiency of yield testing trials and facilitates timely data collection for breeding decisions. Additionally, this approach can be integrated with drone-based phenotyping to further reduce labor and time in breeding programs~\citep{li2024soybeannet}. Besides this, soybean researchers used 3-dimensional imaging technologies such as LiDAR to create new phenotyping tools for soybean breeding purposes~\citep{young2023canopy, young2024soybean}. A similar technique was used for soybean pod detection by employing a depth camera to render a 3-dimensional view of an entire soybean plant~\citep{mathew2023novel}. This method uses the imagery to detect the distance from the plant to the camera, creating a 3-dimensional heat map that is then used to estimate pod count. The methods mentioned above are all based on soybean pod detection or pod counting and use that as a proxy for yield. They have demonstrated significant potential for accurately estimating pod count and, in some cases, providing yield ranking estimates. 

Nevertheless, a better seed yield estimate will be seed count on standing plants. A stronger correlation between soybean yield and the number of seeds (r=0.92) was reported~\citep{wei2020soybean}. Researchers have developed some soybean seed detection models~\citep{uzal2018seed, li2019soybean}. A significant limitation of these works is their reliance on imagery of pods with either black or white backgrounds. They have yet to be implemented in a high-throughput data collection manner in a breeding plot field environment. In these works, they consider seed detection as a common object detection problem that involves locating and identifying each individual separately. Unlike object detection, however, a more specific task called crowd counting aims to directly locate target objects and estimate their count in one shot. Crowd counting approaches often perform much better in densely populated scenes than regular object detection methods. Considering that a soybean field is a crowded scene, researchers from the University of Tokyo treated the seed yield estimation as a crowd-counting problem~\citep{Zhao2023}.  They proposed the P2PNet-Soy model, which was extended from P2PNet~\citep{Song2021}, for soybean seed counting. Key improvements in this model include the integration of k-d tree post-processing, multi-scale multi-reception field feature extraction, and attention mechanisms, significantly reducing the mean absolute error (MAE). The study underscores the importance of considering high-level and low-level features to enhance model accuracy, demonstrating substantial improvements in seed counting and localization performance. However, it is yet needed to address the scale of phenotyping and decision-making in a breeding program that requires high accuracy.

Overall, the main gaps to address in the development of a seamless phenotyping method for seed yield estimation in a breeding program are (a) a large number of plots across different tests, (b) varying genetic variation among plant materials, (c) field environment that is impacted by multiple weather elements complicating data collection and, (d) DL models with improved accuracy and generalizability. Challenges exist in soybean seed detection and quantification with computer vision and machine learning methods. Errors in computer vision tasks can be compounded by background noise, object occlusion, cluttered image environments, variable lighting, and weather conditions, among other factors. In this paper, we propose a method for non-destructively estimating soybean seed yield using computer vision and deep learning. We first capture video data from a ground robot, which will be segmented into individual video frames, i.e. images. We then train a DL model that detects and quantifies seeds captured in these images and uses the estimated seed count to rank plots for breeding decisions. Yield ranking, as opposed to direct computation, provides breeders with an efficient way to compare different experimental lines and candidate varieties, including their relative performance to established checks. Our proposed approach is a great alternative solution to combine yield. It is especially practical and efficient when ground-robot-based imaging with our yield estimation model (P2PNet-Yield) can be used to save time and resources, or when absolute yield measurements are not feasible due to time constraints or machine breakdowns. 

Here are some highlights of our work. First of all, we conducted a 3-year field experiment for data collection, which covers varying genetic variation and field environment. During data collection, we ensured that every side of every detected plant was imaged so seeds occluded by plants or other pods from one side were still captured on the opposite side. Secondly, as for the design of our deep learning architecture, we proposed a yield estimation framework called P2PNet-Yield, which is based on P2PNet-Soy~\citep{Zhao2023} and~\cite{Riera2021}'s two-stage architecture. Thirdly, several strategies were adopted in our model training. For example, to enhance the generalizability of the seed detection model (P2PNet-Soy), we trained it using image data captured under various imaging conditions with different camera sensor effects applied. Last but not least, our method combines high-throughput phenotyping using ground-robot-based imaging for seed detection and quantification. It can efficiently estimate and rank soybean yield for plant breeding decisions to select varieties. The genotype ranking based on seed counting and yield estimation results was conducted to evaluate the performance of our work.

To summarize, the main contributions of our work are (a) the creation of a large-scale image dataset of soybean plants with various genotypic traits collected over three years (2021 to 2023), (b) the development of strategies to improve the accuracy of the seed-counting model trained on fisheye data, including fisheye image correction, data augmentation with random sensor effects, and a spatial adjustment method to account for environmental variations and (c) creation of a yield estimation architecture (P2PNet-Yield) that combines the backbone of the seed counting model (P2PNet-Soy) with a custom yield estimation regressor. 



\section{Materials and Methods}
\label{sec:materials&methods}

\subsection{Field Experiments and Data Collection}
\label{subsec:data}

\subsubsection{Field Experiments}
\label{subsubsec:field_exp}
 
Field experiments consisted of plant breeding trials in the ISU soybean breeding program. We collected data from F$_5$ and F$_7$ filial generations yield plots~\citep{singh2021plant}. Phenotyping was done in the advanced soybean yield trials near Boone, IA (42.020,-93.773, 339 meters above sea level). Robot video data were captured from two fields: (a) in 2021 -  filial generation 5 (F$_5$), number of yield trials = 8500, and (b) in 2023 - filial generation 7 (F$_7$), number of yield trials = 650. Data from the 2021 F$_5$ field was used for training and testing of our seed detection model. In each year, yield trials were in a two-row configuration with a row spacing of 0.76 meters, a seed-to-seed spacing of 3.68 centimeters, and 0.91-meter alleyways. Plot lengths were 2.13 meters in F$_5$ and 5.18 meters in the F$_7$ generation. The plots in these experiments represent a genetically diverse collection of breeding populations representing elite and plant introduction parental stocks~\citep{singh2021plant_ch5}. 

Each year, soybean plots were seeded on a field that had bulk maize (\textit{Zea mays} L.) the previous year. Standard ground preparation methods were practiced. Each field was treated with a post-planting herbicide a month after planting. In addition to chemical weed control, manual weed control was done by routinely walking the plots to remove weeds.


\subsubsection{Data Spatial Adjustment}
\label{subsubsec:spatial_adj}

\begin{figure}[h!tb]
    \centering
    \includegraphics[width=0.7\textwidth]{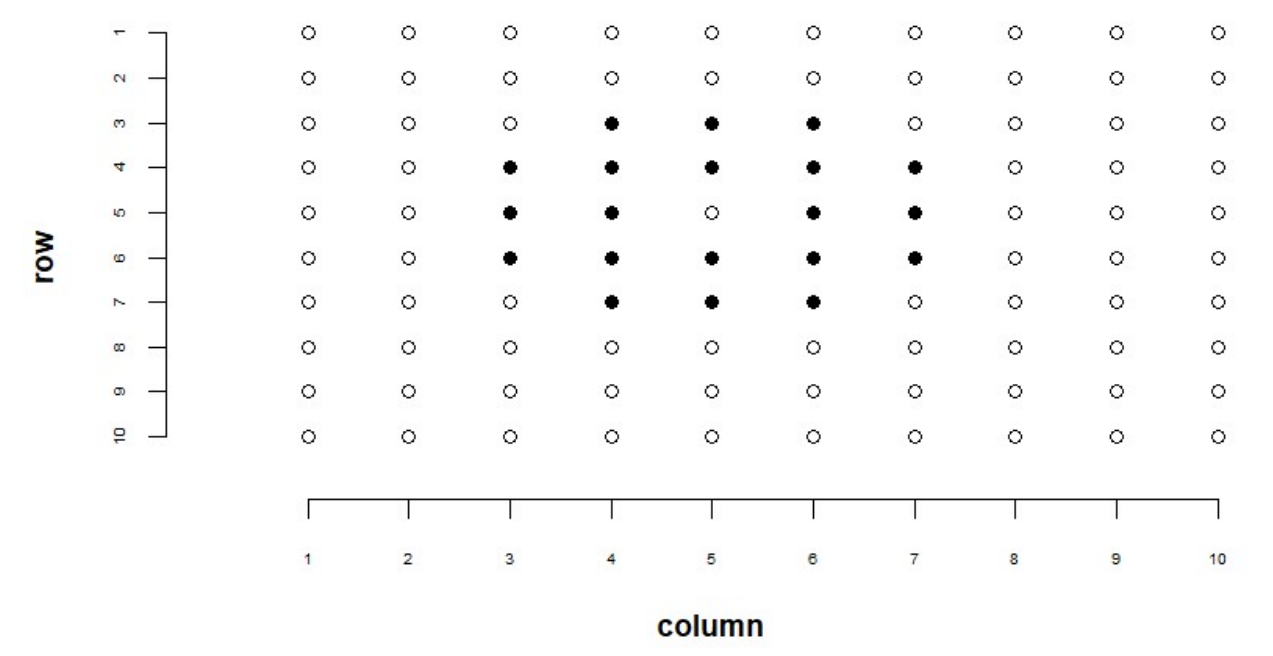}
    \caption{Example of our spatial adjustment grid pattern. Highlighted cells represent plots used in the spatial adjustment. Center cell that is not highlighted represents the cell being adjusted.}
    \label{fig:spatial adjustment grid}
\end{figure}

Spatial adjustment techniques can be applied to breeding plots to help account for environmental variations such as soil properties and reduce non-genetic variability in our data \citep{carroll2024leveraging}. To account for environmental variation in our analysis, we employed the moving grid adjustment method provided by the mvngGrAd package \citep{technow2015r}. This adjustment was performed on our ground truth plot yields, estimated total seed counts, and estimated yields for each plot. Genotype ranking results reported in this paper were conducted on spatially adjusted data. Methods for estimated total seed count and estimated yields are explained in later sections. The spatial adjustment method involves adjusting a plot's value based on the values of its neighboring plots within a defined grid. This grid can be adjusted relative to the plot of interest in the 0, 90, 180, and 270-degree directions based on user definition and need. For our spatial adjustment pattern, we utilized a $5\times5$ grid, excluding the corner plots and the center plot. This grid configuration is depicted in Figure~\ref{fig:spatial adjustment grid}. The \texttt{movingGrid()} function, part of the \texttt{mvngGrAd} package, performs the spatial adjustment by calculating moving means for each plot based on this grid of neighboring plots. The adjusted phenotypic value $p_{i,\text{adj}}$ is calculated using the following formula:

\begin{equation} \label{eq:p_(i,adj)}
p_{i,\text{adj}} = p_{i,\text{obs}} - b(x_i - \bar{x})
\end{equation}

where:
\begin{itemize}
    \item $ p_{i,\text{adj}} $ is the adjusted phenotypic value for entry $ i $.
    \item $ p_{i,\text{obs}} $ is the observed phenotypic value for entry $ i $.
    \item $ b $ is the coefficient representing the relationship between the growing conditions and the observed phenotypic value.
    \item $ x_i $ is the moving mean phenotypic value for entry $ i $, calculated as the mean of the cells within the grid around entry $ i $.
    \item $ \bar{x} $ is the overall mean of the moving means $ x_i $.
\end{itemize}


\subsubsection{Ground Robot Based Image Data Collection}
\label{subsubsec:data_collect}

The robot used in this experiment was the Terrasentia, developed by Earthsense (Earthsense, Champaign, USA)~\citep{mcguire2021}. This robotic platform is equipped with two side-facing cameras, a forward-facing camera, an upward-facing camera, vertical and horizontal LiDAR sensors, and an RTK GPS. Figure~\ref{fig:data_collection_pipline} shows the robot operating in a field. The video cameras on this robot were fitted with fisheye lenses, ensuring comprehensive capture of the soybean plants from the base to the top. For this experiment, the two side-facing cameras were utilized. Data were collected as side-view videos of soybean plots after the entire field had reached full physiological maturity (stage R8)~\citep{fehr1971stage}. Video data was collected by manually navigating the robot between each row of soybeans (Figure~\ref{fig:plotCollection}). Videos were recorded at a resolution of $1920\times1080$ pixels. Each traverse of the robot through the field resulted in a continuous video recording, which we will refer to henceforth as a "collection." Each collection began at the start of a pass and ended once the robot reached the end of that pass. The robot was maneuvered in a serpentine pattern to ensure imaging of both sides of every row in each plot. This approach aimed to capture pods that might have been obstructed from one side but were visible from the other. The dual side-mounted cameras on the Terrasentia allowed us to achieve this with a minimum number of collections. Although a single collection captures multiple plots, three collections were needed to fully capture any one plot. 

\subsubsection{Data Preprocessing}
\label{subsubsec:preprocessing}

\begin{figure}[h!tb]
    \centering
    \includegraphics[width=1\textwidth]{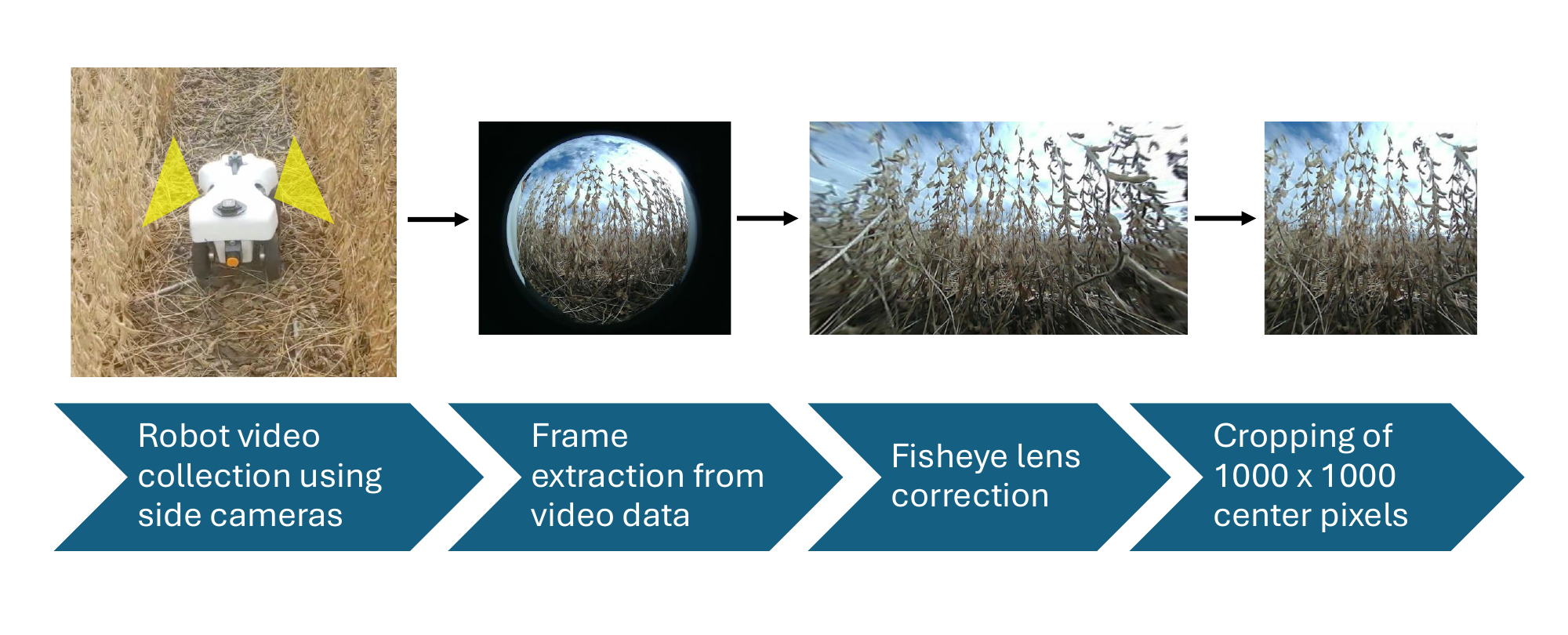} 
    \caption{This figure demonstrates the pipeline from in-field data collection and the post-processing needed for use in our P2PNet-Yield model. The first figure shows our Terrasentia robot operating in a mature soybean field. As the robot moves through the field, the two side-mounted cameras collect fisheye video data. Individual frames are then extracted, corrected for fisheye distortion, and cropped to remove blurry edges.}
    \label{fig:data_collection_pipline}
\end{figure}

Following the collection of video data, individual frames were extracted from each video using Python~\citep{python} scripts. As previously mentioned, these frames were captured using fisheye lenses, which introduced distortion artifacts. To achieve accurate seed counts and yield estimations, it was essential to remove these distortions. An OpenCV~\citep{opencv_library} tool was employed to calibrate the images, correcting for fisheye lens distortion. Calibration was conducted using a standard checkerboard pattern, with multiple images captured from varying angles and distances to ensure precise calibration. This was done in accordance with a pre-existing method \citep{jiang2017fisheye}. Through this calibration process, the intrinsic camera parameters of the fisheye camera were obtained: the focal length $(f_x,f_y)$ in pixels is $(410,410)$, and the principal point $(p_x,p_y)$ in pixels is $(383,526)$. Using these parameters, the fisheye distortion was effectively corrected (see Figure~\ref{fig:data_collection_pipline}). After correcting the fisheye distortion, the edges along the sides, as well as the top and bottom of the images, were left blurry and sometimes still distorted. A central area measuring $1000\times1000$ pixels was then cropped. It was later utilized for image annotation, training, and total seed count (TSC) estimations. This ensured that the outside blurred regions were removed, mitigating the background noise (see Figure~\ref{fig:data_collection_pipline}). 

Each calibrated frame was then assigned to its respective plot using the on-board vertical LiDAR sensor to detect the start and stop points of each plot. These start and stop points were generated by Earthsense's proprietary data processing tools. They were provided as a CSV file containing a series of time points representing the start and stop times for each plot. Each frame in the video was associated with a corresponding time point. Python scripts were used to extract these time points and their associated frames, organizing each set of images into their respective plots. We manually proofed the accuracy of the plot segmentation data provided by Earthsense by randomly checking about ~50\% of the images, and no discrepancies were found.

Once organized by plot, the images were further sorted by their respective row and side assignments. Specifically, the plot images were organized first by plot name (range, pass), then by row number (1 or 2), and by row side (A or B). This resulted in four sets of images, each set containing approximately 100 images, or around 400 images per plot. In total, 338,793 images were generated from the F$_7$ generation material in 2023. Plots from 2023 were used to test the accuracy of the seed count and yield estimation model in ranking genotype yield. 

Given that the number of images per plot could be large, a selection process was employed for seed counting and yield estimation. As illustrated in Figure~\ref{fig:sampling}, each row was equidistantly divided into eight sections using seven splitters, with images from the middle five splitters chosen for analysis. The two rows of the same plot were treated as a single row, resulting in ten images per side and twenty images per plot. These twenty sample images were evenly distributed across the plot, containing the most representative information. The first and last splitters were excluded from the analysis, as they were too close to the start or end of each row and occasionally did not contain any plants.

\subsubsection{Ground Truth Seed Count and Yield Data Collection}
\label{subsubsec:ground_truth}

From the set of images captured from the F$_5$ generation material in 2021, a subset of 1,200 images was randomly selected for seed annotation. Expert raters conducted these annotations, marking only visible soybean seeds in each image using point annotations. The annotations were facilitated by the Label Studio software \citep{labelstudio}. Only seeds that were clearly discernible by human raters were annotated. Seeds that were too indistinct to separate and the ones located in background plots were excluded (see Figure~\ref{fig:seed_annotations}). This subset of images was taken at any time of day from 8:00 a.m. to 6:00 p.m. It represents a range of weather and lighting conditions, as well as varying levels of soybean lodging and occlusion. The genotypes captured in these images exhibit diverse pubescence and pod wall colors. The images were selected across the plot, ensuring the inclusion of both seed-rich and seed-scarce images. This strategy enhances the model's ability to detect seeds under a wide variety of conditions, thereby improving its generalizability.

Another set of ground truth data is the ground truth yield. Ground truth yield data for each plot was obtained using either a Zurn (Zürn Harvesting, Schöntal-Westernhausen, Baden-Württemberg, Germany) or Almaco (Almaco, Nevada, IA, USA) plot combine, which provided seed yield measurements in kilograms for each plot. Ground truth yield data were collected for all plots imaged by the Terrasentia (See section \ref{subsubsec:data_collect}). To ensure consistency in yield measurements across fields with varying plot sizes, yields were adjusted to 13\% moisture and converted to metric tons/hectare (t/ha) for each harvested plot. 

\subsection{Seed Counting on Fisheye Data}
\label{subsec:seed_counting}


The undistortion process applied to the fisheye images resulted in more consistent patterns of soybean seeds. This consistency made it easier to train the feature extraction backbone of the seed counting model, P2PNet-Soy~\citep{Zhao2023}. The model’s pre-trained weights were utilized to further enhance training efficiency. We trained the P2PNet-Soy model on our corrected image data to help it learn feature maps that contain relevant information about the seeds. 

To prepare the training datasets, we augmented our original corrected image dataset by applying random camera sensor effects, such as noise, blurring, chromatic aberration, and exposure adjustment~\citep{Carlson2018}. The purpose of this data augmentation strategy was to reduce the differences between images captured by different cameras in varying environments. For instance, noise was added to simulate the common artifacts found in low-quality images due to sensor limitations. Blurring could reduce edge sharpness to account for motion or focus issues. Chromatic aberration was introduced to replicate the color fringing caused by lens imperfections, and exposure adjustments helped simulate changes in lighting conditions or camera quality.

The data augmentation process started with our original dataset. With data augmentation, we obtained the augmented datasets, which include both the original images and those modified by these camera sensor effects. Each original image has one augmented version with various sensor effects applied. Instead of uniformly applying these effects to all images, we introduced variability by randomly adjusting the intensities of each effect. These augmentations were implemented using a tool developed by~\cite{Carlson2018}, which can automatically select random parameter settings for each effect. By incorporating these augmentations, we aimed to make our model more robust to variations in image quality across different cameras and environments. The improvements brought about by this augmentation technique are discussed in Section~\ref{subsec:sc_results}.

With the augmented data, we prepared three datasets for training: UTokyo, ISU2021, and ISU2021-AUG. The UTokyo dataset includes 126 images for training and 27 for evaluation, which was provided by the University of Tokyo team who introduced P2PNet-Soy~\citep{Zhao2023}. The ISU2021 dataset comprises 1,200 annotated images from the 2021 F$_5$ field, divided into two subsets: 1,007 images for training and 193 for evaluation. The ISU2021-AUG dataset was created by applying camera sensor effects to our ISU2021 dataset and consists of augmented versions of the 1,007 training images from ISU2021. 

We trained the seed counting model on four various combinations of these three datasets to identify the best model for seed detection and counting. These four combinations were \texttt{ISU\_NO\_AUG} (using only the ISU2021 data), \texttt{MIX\_NO\_AUG} (using ISU2021 and UTokyo), \texttt{ISU\_AUG} (using ISU2021 and ISU2021-Aug) and \texttt{MIX\_AUG} (using ISU2021, ISU2021-Aug, UTokyo). During training, the weights from the original P2PNet-Soy model were employed, and most hyperparameters were kept at their default settings. For each combination, the model was trained for 100 epochs. The model was validated using the mean squared error (MSE), mean absolute error (MAE), and mean absolute percentage error (MAPE), which can be found in Section~\ref{subsec:sc_results}. 



\subsection{Yield Estimation Architecture}
\label{subsec:yield_estimation}

Inspired by previous works~\citep{Riera2021}, our DL architecture for yield estimation, named P2PNet-Yield, consists of two modules: the Feature Extraction Module and the Yield Regression Module. The architecture is depicted in Figure~\ref{fig:yield_architecture}. The backbone of the seed counting model (P2PNet-Soy) serves as the Feature Extraction Module. The feature map extracted by this module contains valuable information that can be used for seed detection. For each plot, we applied our Feature Extraction Module to the 20 sample images selected (as discussed in Section~\ref{subsubsec:preprocessing}), obtaining 20 feature maps.

Once these feature maps were obtained, the next task was to predict yield using these feature maps. To fuse information from these feature maps, we summed the 10 feature maps from the same side and then concatenated them. The fused feature map was then fed into the Yield Regression Module. The Yield Regression Module consists of one convolution layer (conv) followed by a max-pooling layer, which is then flattened and followed by three fully connected layers (fc1, fc2, and fc3). The output of the Yield Regression Module is the yield (in t/ha) for the plot. During training, the layers in the Feature Extraction Module were frozen. The batch size was set to 8, and the models were trained for 50 epochs. We adopted the Adam optimizer~\citep{Adam2014} and used element-wise mean squared error as our loss function.

\section{Results}
\label{sec:results}

The training and testing experiments of our deep learning architecture utilized the PyTorch library version 1.8.0 with CUDA 11.1 support for an NVIDIA GPU. For all the results presented in this section, we used an NVIDIA A100 GPU with 80GB VRAM running on a CPU with Intel Skylake Xeon processors with 512GB RAM.

\subsection{Seed Counting}
\label{subsec:sc_results}


In our experiments, we noticed that the seed counting model (P2PNet-Soy) trained on our corrected fisheye images without any data augmentation did not generalize well and showed overcounting issues during testing (as shown in Figure~\ref{fig:sc_correlation_GTvsP_1}). Specifically, the P2PNet-Soy model struggled to distinguish between foreground and background soybean plants, leading to persistent seed detection in the background. This was likely due to differences in cameras and imaging conditions across the test datasets, which included images randomly selected from both our datasets and the one from the P2PNet-Soy team.

With data augmentation utilizing camera sensor effects, the performance of the seed counting model is improved. As discussed in~\ref{subsec:seed_counting}, Mean squared error (MSE), mean absolute error (MAE), and mean absolute percentage error (MAPE) were calculated to evaluate the fine-tuned P2PNet-Soy model trained on different combinations of the training datasets. For all three metrics, augmentation improved the model performance, i.e., \texttt{ISU\_AUG} was better than \texttt{ISU\_NO\_AUG}, and \texttt{MIX\_AUG} was better than \texttt{MIX\_NO\_AUG} (Table~\ref{tab:SC_test_accuracy}). Larger datasets with more environmental variations performed better than smaller data, i.e., \texttt{MIX\_NO\_AUG} was better than \texttt{ISU\_NO\_AUG}, and \texttt{MIX\_AUG} was better than \texttt{ISU\_AUG} (Table~\ref{tab:SC_test_accuracy}). The MSE, MAE, and MAPE were lowest for \texttt{MIX\_AUG}. In the meanwhile, \texttt{MIX\_NO\_AUG} and \texttt{ISU\_AUG} had a similar performance with slightly better MSE and MAE for \texttt{ISU\_NO\_AUG} (Table~\ref{tab:SC_test_accuracy}). 

These results are visualized through correlation plots that plot the ground truth and estimated counts for each of the four dataset combinations (Figure~\ref{fig:sc_correlation_GTvsP}). The correlations from \texttt{ISU\_NO\_AUG} showed an upwards bias in the estimated counts, particularly at higher seed count values ($R^2$ value of 0.80, Figure~\ref{fig:sc_correlation_GTvsP_1}). The \texttt{ISU\_AUG} had less bias, although still trending of overestimating at higher seed count values, but didn't show a tighter fit to the regression line ($R^2$ value of 0.87, Figure~\ref{fig:sc_correlation_GTvsP_3}). The \texttt{MIX\_NO\_AUG} tends to overestimate at lower values ($R^2$ value of 0.77, Figure~\ref{fig:sc_correlation_GTvsP_2}). The best fit was noted for \texttt{MIX\_AUG} dataset combination with the highest $R^2$ ($R^2$ value of 0.87, Figure~\ref{fig:sc_correlation_GTvsP_4}). The plots of residuals with ground truth show that the model trained on \texttt{MIX\_AUG}, i.e., mixed datasets with data augmentation, performs the best. The residuals were distributed in a narrower band around 0 and did not show any deviation (Figure~\ref{fig:sc_residual}).

The results indicate that the model trained on mixed datasets with data augmentation performs best as per MSE, MAE, and MAPE, correlations, and residual plots. Data augmentation using camera sensor effects effectively reduced over-counting. Errors were further minimized by combining data from different sources.

\subsection{Applications in Plant Breeding and Selection}
\label{subsec:application in plant breeding}



The application of P2PNet-Soy \citep{Zhao2023} and our P2PNet-Yield model was tested in two scenarios to demonstrate the usefulness in a variety development plant breeding program. Both models used the weights trained on the \texttt{MIX\_AUG} dataset combination. In the first scenario, we used the P2PNet-Soy model for seed counting to assign ranks for experimental lines, followed by breeding selection decisions. In the second scenario, we used the P2PNet-Yield model to estimate seed yield (t/ha), assigned ranks for the experimental lines to make breeding selection decisions. We report an $R^2$ value of 0.06 between estimated TSC values from P2PNet-Soy and estimated yield from P2PNet-Yield (Figure \ref{fig:yield_vs_TSC}). We evaluated three essential selection metrics — accuracy, sensitivity, and specificity — using selection thresholds of 30\%, 20\%, and 10\%. These metrics were calculated using spatially adjusted ground truth plot yields and spatially adjusted total seed counts (TSC) from the 2023 F$_7$ material, as these advanced yield tests were grown with two replications. We also performed these analyses on raw ground truth yield (i.e., combine harvestable seed yield) and spatially unadjusted TSC values.


\subsubsection{Variety Ranking and Selections Based on Seed Counting using P2PNet-Soy}
\label{subsec:ranking_sc_results}


Accuracy and specificity scores were relatively high across all three selection thresholds. We note that when the selection threshold becomes more stringent, accuracy and specificity scores increase, while the inverse is true for sensitivity. At 10\%, 20\%, and 30\% selection cut-off, the accuracy values were 0.86, 0.76, and 0.70, respectively. Similarly, the specificity values were highest at a more stringent cut-off (0.92 at 10\%, 0.85 at 20\%, and 0.78 at 30\%). In contrast, sensitivity scores progressively increased as the selection cut-offs were increased, i.e., 0.31 at 10\%, 0.40 at 20\%, and 0.50 at 30\% (Figure~\ref{fig:acc_sens_spec_1}). The number of observations for TP and TN was highest at the 10\% cut-off, and the total number of correct classes (i.e., TP and TN) reduced at 20\% and were lowest at the 30\% (Table~\ref{tab:tp_tn_fp_fn}, Figure~\ref{fig:venn diagram}). The FP and FN values were nearly identical within each selection threshold and increased from 10\% to 30\%.

\subsubsection{Variety Ranking and Selections Based on Yield Estimation using P2PNet-Yield}
\label{subsec:yield_results}

In addition to the analysis based on estimated TSC, we also evaluated genotype ranking based on estimated yield. When using our proposed P2PNet-Yield model, we note that as the selection threshold becomes more stringent, accuracy and specificity scores increase while the inverse is true for sensitivity. Accuracy and specificity scores were high across all three selection thresholds. At 10\%, 20\%, and 30\% selection cut-off, the accuracy values were 0.83, 0.70, and 0.60, respectively. Similarly, the specificity values were highest at a more stringent cut-off (0.91 at 10\%, 0.81 at 20\%, and 0.71 at 30\%). In contrast, sensitivity scores progressively increased as the selection cut-offs were increased, i.e., 0.17 at 10\%, 0.25 at 20\%, and 0.33 at 30\% (Figure~\ref{fig:acc_sens_spec_2}). The number of observations for TP and TN was highest at the 10\% cut-off, and the total number of correct classes (i.e., TP and TN) reduced at 20\% and were lowest at the 30\% (Table~\ref{tab:tp_tn_fp_fn}). The FP and FN values were nearly identical within each selection threshold and increased from 10\% to 30\%. Overall, trends from P2PNet-Yield results were similar to P2PNet-Soy.

\subsection{P2PNet-Yield Performance Under Optimal Conditions}
\label{subsec:p2pnet-yield results}

To assess the P2PNet-Yield model's performance, we manually curated a subset of 200 plots from the 2023 F$_7$ dataset which included 650 plots, i.e., we used ~30\% plots for this analysis. These plots did not have any anomalies, such as severe lodging, severe disease, etc. During planting, harvest, and intermediate periods, notes were recorded for plots exhibiting unique issues or characteristics, such as disease, lodging, or large gaps ($>0.5$m gaps) of missing plants in the harvested rows. We also excluded plots with bad imaging caused by overexposure or camera misplacement. This resulted in a refined dataset comprising 100 plots for training and 100 plots for testing for our P2PNet-Yield model. The $R^2$ was 0.38 between the estimated yield and ground truth yield for these plots and an MSE of 6.53. No correlation was noted in the uncurated dataset (Figure~\ref{fig:yield_result}). Although a manual selection was made to select high-quality plots from the dataset, these results demonstrate the potential efficacy of this architecture in yield estimation in high-quality field experiments.

\section{Discussion}
\label{sec:discussion}

Much of the related research in yield or yield-related trait estimation relies on controlled imaging environments to estimate pod or seed counts \citep{li2019soybean, uzal2018seed, zhang2023high}. These studies typically use imagery of harvested soybean pods against a uniform white or black background to train detection and quantification models. This approach simplifies seed detection by eliminating background noise and ensuring high image quality. Other research expands on this by imaging entire soybean plants post-maturity \citep{xiang2023yolo, yu2024accurate}. In these cases, the entire plant is captured against a black background, allowing for pod detection across the whole plant.

We proposed an innovative approach for estimating soybean seed yield using fisheye imagery data to detect seeds and estimate yield. Fisheye images, while providing comprehensive plant information, pose significant challenges due to distortion. To overcome this, we calibrated the fisheye camera to correct the images. Besides this, we improved the diversity of imaging conditions through data augmentation with camera sensor effects to enhance the generalization of the seed counting model. Our experiments with various dataset combinations revealed that models trained on mixed datasets with data augmentation yielded the best performance. 

We designed an architecture integrating a Feature Extraction Module and a Yield Regression Module, demonstrating satisfactory yield estimation. Our results on seed counts and seed yield per area are similar to those previously achieved with pod counting using a ground robot \citep{Riera2021}. Our genotype ranking analysis indicates that both estimated total seed counts (TSC) and estimated yield are effective for down-selecting poorly performing lines, which is similar to what was previously reported with image-based pod counting \citep{Riera2021}. Although both P2PNet-Soy and P2PNet-Yield methods exhibit low sensitivity scores — suggesting that both methods are sub-optimal for identifying top-performing lines — high specificity scores confirm their utility in eliminating the poor-performing lines. This is particularly useful in early-stage yield trials. While TSC as a selection method performs marginally better at all three selection metrics than estimated yield (t/ha) using the same dataset, our P2PNet-Yield model, when trained and tested on high-quality plots, exhibits a higher correlation between estimated and actual yields. We report the reasonable capability of our P2PNet-Yield model to estimate yield similar to the plot combine yield, at least in ranks, which is useful for a plant breeder to make selection decisions. These results also highlight the importance of high-quality data for optimal performance. 


Field-based studies often utilize small plots and relatively small datasets. Researchers have employed ground robots to extend this approach to larger plots \citep{Riera2021, mcguire2021}. These experiments focused on detecting soybean pods. Our research builds upon these efforts by introducing a novel pipeline for seed detection in a variety of development programs with field-based experiments, moving beyond controlled environment settings. While research has been done to present seed counting methods for yield estimation in a field environment, it was built on a small dataset (24 accessions; 374 images of individual plants) and lacked a high-throughput data collection method \citep{Zhao2023}. Our work is the first to implement seed counting as a high-throughput method for yield estimation in non-controlled environments and to analyze full-sized breeding plots for cultivar development purposes.

However, our work has several limitations and areas for improvement. First, the yield estimation model is highly dependent on image quality. The accuracy of yield estimation deteriorates when parts of the soybean plants are obscured due to poor lighting or occlusion. This is because the Feature Extraction Module may struggle to extract good feature maps with seed information in these challenging areas. Second, the current methods for sampling and feature fusion are based on experience and could be optimized. To be more specific, the number of splitters in data sampling (currently 20 per plot) and how we combine the 20 feature maps of sample images for each plot is worth further exploration. More effective approaches for sample selection and feature fusion could enhance model performance. Additionally, our use of fisheye lenses for comprehensive plant capture introduces challenges. The use of fisheye lenses, even after correction, leaves our images with distorted and blurry edges, requiring cropping to remove these regions. This process results in data loss from the very top and bottom of the images, limiting the model’s ability to detect seeds in these regions. Using a higher-resolution fisheye camera system would eliminate these blurry edges and would no longer require cropping and, therefore, no data loss. This could increase estimated total seed count (TSC) and yield (t/ha) accuracy, improving yield ranking.


To improve our P2PNet-Yield model, additional factors such as seed size and weight could be incorporated. Researchers have developed a transfer learning approach to automatically detect seed size in controlled imaging environments~\citep{yang2021high}. Other researchers have expanded on this research by using imagery of soybean pods to detect pod width and length \citep{yang2022synthetic, ning2022extraction}. Including these additional data points would provide a more comprehensive yield estimation model, improving the accuracy of selection decisions by capturing key phenotypic traits that influence overall yield. Utilizing additional data points, such as vegetative indices from hyperspectral imagery and soil conditions, could also provide additional improvements in a yield estimation model~\citep{gupta2023agri, chattopadhyay2023comprehensive}. 

Unmanned Aerial Vehicles (UAVs) also play a vital role in modern yield prediction for breeding plots. Due to their rapid field-sensing capabilities, UAVs can survey numerous plots within minutes, making them an efficient tool for in-season data collection. Previous studies have demonstrated the effectiveness of using vegetative indices~\citep{maimaitijiang2020soybean}, canopy texture data~\citep{alabi2022estimation}, and canopy area~\citep{yu2016development} as inputs for machine learning-based soybean seed yield prediction models. Integrating in-season UAV scouting with our late-season seed detection and yield estimation model could potentially enhance yield prediction accuracy. However, UAVs often operate at heights over 30 meters, which limits the visibility of lower plant regions and lacks the ability to sense fine details, such as individual seeds within pods, due to canopy overlap. Recent advancements, using UAV imagery angled at 53-58 degrees from approximately 4 meters, have shown promise in detecting soybean pods~\citep{li2024soybeannet}. This suggests that future improvements in camera technology could make UAV platforms compatible with high-resolution models like P2PNet-Yield thereby decreasing the time needed to scout a field.

Operating at a speed of 4 kph, the Terrasentia robot has a battery life of approximately 2.5 to 3 hours. With a 2TB onboard storage capacity, the robot is capable of storing about 12 hours of collected data before requiring offloading to an external storage device. Tasks such as recharging, battery swapping, and data offloading add time to the data collection process and limit the number of plots an operator can image per session. To streamline and integrate these processes, our proposed data collection pipeline could be embedded within a farm-wide data network, utilizing edge and cloud computing to automate data offloading and potentially enable real-time yield estimation~\citep{singh2024smart}. Citizen science networks have already shown the effectiveness of collaborative efforts in enhancing crop management and plant health~\citep{chiranjeevi2023deep}. Establishing farm networks can provide farmers with comprehensive information on optimal crop management practices. Deploying our model in a networked setting would allow it to be trained across more diverse genetic material, thus enhancing both the generalizability and accuracy of our model. By incorporating a wide range of genetic diversity and deploying our model across multiple environments, our yield estimation framework could significantly contribute to increased genetic gain~\citep{krause2023models}.

Previous research has utilized depth cameras to detect soybean pods~\citep{mathew2023novel}. A promising direction for future work would be to implement this approach as part of a two-stage system: the depth camera could first locate pods, and then our P2PNet-yield model could focus specifically on these areas to identify seeds within the pods more accurately. Similar 3D imaging techniques, such as LiDAR, have been used to create full 3D reconstructions of soybean plants~\citep{young2024soybean, young2023canopy}. Integrating our yield estimation model with these 3D plant structures could support breeders in developing enhanced soybean varieties, enabling the creation of an optimized crop ideotype for breeding~\citep{singh2021plant_ch25}.

\section{Conclusion}
\label{sec:conclusion}

Traditional yield data collection methods are often costly, time-consuming, and susceptible to equipment malfunctions and maintenance issues. Our research introduces an innovative pipeline that streamlines the yield collection process, offering a more efficient and cost-effective solution for yield estimation and analysis. Specifically, we propose the P2PNet-Yield for seed yield estimation in soybean plots. We developed an end-to-end pipeline that demonstrates the effectiveness of ground robots for high-throughput soybean imaging, seed counting, and yield estimation. Additionally, we illustrated the application of seed counts and yield estimation for effective variety selection, particularly in discarding poor-performing varieties. In the future, as these ground-robot-based imaging platforms and DL methods continually improve, plant breeders will be able to forego combine harvesting at some locations and instead use P2PNet-Yield or related systems to obtain seed yield data.

\section*{Conflict of Interest Statement}

The authors declare that the research was conducted in the absence of any commercial or financial relationships that could be construed as a potential conflict of interest.

\section*{Author Contributions}


Jiale Feng: Conceptualization, Methodology, Software, Formal Analysis, Investigation, Data Curation, Visualization, Writing - Original Draft, Writing - Review \& Editing

Samuel W. Blair: Conceptualization, Methodology, Software, Formal Analysis, Investigation, Data Curation, Visualization, Writing - Original Draft, Writing - Review \& Editing

Timilehin Ayanlade: Software, Data Curation, Writing - Review \& Editing

Aditya Balu: Software, Validation, Formal analysis, Writing - Review \& Editing, 

Baskar Ganapathysubramanian: Resources, Writing - Review \& Editing, Funding acquisition

Arti Singh: Resources, Writing - Review \& Editing, Funding acquisition

Soumik Sarkar: Conceptualization, Methodology, Resources, Writing - Review \& Editing, Supervision, Funding acquisition

Asheesh K. Singh: Conceptualization, Methodology, Resources, Writing - Original Draft, Writing - Review \& Editing, Supervision, Project administration, Funding acquisition

Jiale Feng and Samuel W. Blair are co-first authors, and contributed equally to this work.


\section*{Funding}
Iowa Soybean Association; USDA CRIS project IOW04714; AI Institute for Resilient Agriculture, Grant/Award Number:
USDA-NIFA \#2021-67021-35329; COALESCE: COntext Aware LEarning for Sustainable CybEr-Agricultural Systems,
Grant/Award Number: CPS Frontier \# 1954556; Smart Integrated Farm Network for Rural Agricultural Communities
(SIRAC), Grant/Award Number: NSF S\&CC \#1952045; RF Baker Center for Plant Breeding; Plant Sciences Institute; G.F. Sprague Chair in Agronomy.

\section*{Acknowledgments}
The authors are thankful to Brian Scott, Jennifer Hicks, Ryan Dunn and David Zimmerman for their efforts in field experiments. We also thank the many graduate and undergraduate students who assisted in data collection. 

\section*{Data Availability Statement}
The datasets generated for this study can be obtained by emailing A.K. Singh.

\bibliographystyle{Frontiers-Harvard} 
\bibliography{FPS2024}

\begin{thebibliography}{49}
\providecommand{\natexlab}[1]{#1}
\expandafter\ifx\csname urlstyle\endcsname\relax
  \providecommand{\doi}[1]{doi:\discretionary{}{}{}#1}\else
  \providecommand{\doi}{doi:\discretionary{}{}{}\begingroup \urlstyle{rm}\Url}\fi
\providecommand{\selectlanguage}[1]{\relax}
\providecommand{\bibAnnoteFile}[1]{%
  \IfFileExists{#1}{\begin{quotation}\noindent\textsc{Key:} #1\\
  \textsc{Annotation:}\ \input{#1}\end{quotation}}{}}
\providecommand{\bibAnnote}[2]{%
  \begin{quotation}\noindent\textsc{Key:} #1\\
  \textsc{Annotation:}\ #2\end{quotation}}

\bibitem[{Alabi et~al.(2022)Alabi, Abebe, Chigeza, and Fowobaje}]{alabi2022estimation}
Alabi, T.~R., Abebe, A.~T., Chigeza, G., and Fowobaje, K.~R. (2022).
\newblock Estimation of soybean grain yield from multispectral high-resolution uav data with machine learning models in west africa.
\newblock \emph{Remote Sensing Applications: Society and Environment} 27, 100782
\bibAnnoteFile{alabi2022estimation}

\bibitem[{Bargoti and Underwood(2017)}]{bargoti2017image}
Bargoti, S. and Underwood, J.~P. (2017).
\newblock Image segmentation for fruit detection and yield estimation in apple orchards.
\newblock \emph{Journal of Field Robotics} 34, 1039--1060
\bibAnnoteFile{bargoti2017image}

\bibitem[{Bradski et~al.(2024)Bradski, Kaehler et~al.}]{opencv_library}
Bradski, G., Kaehler, A., et~al. (2024).
\newblock \emph{OpenCV: Open Source Computer Vision Library}.
\newblock Version 4.8.0
\bibAnnoteFile{opencv_library}

\bibitem[{Carlson et~al.(2018)Carlson, Skinner, Vasudevan, and Johnson-Roberson}]{Carlson2018}
Carlson, A., Skinner, K.~A., Vasudevan, R., and Johnson-Roberson, M. (2018).
\newblock Modeling camera effects to improve visual learning from synthetic data.
\newblock \emph{Lecture Notes in Computer Science (including subseries Lecture Notes in Artificial Intelligence and Lecture Notes in Bioinformatics)} 11129 LNCS, 505--520
\bibAnnoteFile{Carlson2018}

\bibitem[{Carroll et~al.(2024)Carroll, Riera, Miller, Dixon, Ganapathysubramanian, Sarkar et~al.}]{carroll2024leveraging}
Carroll, M.~E., Riera, L.~G., Miller, B.~A., Dixon, P.~M., Ganapathysubramanian, B., Sarkar, S., et~al. (2024).
\newblock Leveraging soil mapping and machine learning to improve spatial adjustments in plant breeding trials.
\newblock \emph{Crop Science}
\bibAnnoteFile{carroll2024leveraging}

\bibitem[{Chattopadhyay et~al.(2023)Chattopadhyay, Gupta, Carroll, Raigne, Ganapathysubramanian, Singh et~al.}]{chattopadhyay2023comprehensive}
Chattopadhyay, S., Gupta, A., Carroll, M., Raigne, J., Ganapathysubramanian, B., Singh, A., et~al. (2023).
\newblock A comprehensive study on soybean yield prediction using soil and hyperspectral reflectance data.
\newblock (Preprints)
\bibAnnoteFile{chattopadhyay2023comprehensive}

\bibitem[{Chiranjeevi et~al.(2023)Chiranjeevi, Sadaati, Deng, Koushik, Jubery, Mueller et~al.}]{chiranjeevi2023deep}
Chiranjeevi, S., Sadaati, M., Deng, Z.~K., Koushik, J., Jubery, T.~Z., Mueller, D., et~al. (2023).
\newblock Deep learning powered real-time identification of insects using citizen science data.
\newblock \emph{arXiv preprint arXiv:2306.02507}
\bibAnnoteFile{chiranjeevi2023deep}

\bibitem[{Fehr et~al.(1971)Fehr, Caviness, Burmood, and Pennington}]{fehr1971stage}
Fehr, W., Caviness, C., Burmood, D., and Pennington, J. (1971).
\newblock Stage of development descriptions for soybeans, glycine max (l.) merrill 1.
\newblock \emph{Crop Science} 11, 929--931
\bibAnnoteFile{fehr1971stage}

\bibitem[{Grimm et~al.(2019)Grimm, Herzog, Rist, Kicherer, Toepfer, and Steinhage}]{grimm2019adaptable}
Grimm, J., Herzog, K., Rist, F., Kicherer, A., Toepfer, R., and Steinhage, V. (2019).
\newblock An adaptable approach to automated visual detection of plant organs with applications in grapevine breeding.
\newblock \emph{Biosystems Engineering} 183, 170--183
\bibAnnoteFile{grimm2019adaptable}

\bibitem[{Guadagna et~al.(2023)Guadagna, Fernandes, Chen, Santamaria, Teng, Frioni et~al.}]{guadagna2023using}
Guadagna, P., Fernandes, M., Chen, F., Santamaria, A., Teng, T., Frioni, T., et~al. (2023).
\newblock Using deep learning for pruning region detection and plant organ segmentation in dormant spur-pruned grapevines.
\newblock \emph{Precision Agriculture} 24, 1547--1569
\bibAnnoteFile{guadagna2023using}

\bibitem[{Gupta and Singh(2023)}]{gupta2023agri}
Gupta, A. and Singh, A. (2023).
\newblock Agri-gnn: A novel genotypic-topological graph neural network framework built on graphsage for optimized yield prediction.
\newblock \emph{arXiv preprint arXiv:2310.13037}
\bibAnnoteFile{gupta2023agri}

\bibitem[{He et~al.(2023)He, Ma, Guan, Wang, and Shen}]{he2023recognition}
He, H., Ma, X., Guan, H., Wang, F., and Shen, P. (2023).
\newblock Recognition of soybean pods and yield prediction based on improved deep learning model.
\newblock \emph{Frontiers in Plant Science} 13, 1096619
\bibAnnoteFile{he2023recognition}

\bibitem[{Herr et~al.(2023)Herr, Adak, Carroll, Elango, Kar, Li et~al.}]{herr2023unoccupied}
Herr, A.~W., Adak, A., Carroll, M.~E., Elango, D., Kar, S., Li, C., et~al. (2023).
\newblock Unoccupied aerial systems imagery for phenotyping in cotton, maize, soybean, and wheat breeding.
\newblock \emph{Crop Science} 63, 1722--1749
\bibAnnoteFile{herr2023unoccupied}

\bibitem[{{HumanSignal, Inc.}(2023)}]{labelstudio}
[Dataset] {HumanSignal, Inc.} (2023).
\newblock {Label Studio} (version 1.13.0).
\newblock Open Source Data Labeling Platform
\bibAnnoteFile{labelstudio}

\bibitem[{Jiang(2017)}]{jiang2017fisheye}
[Dataset] Jiang, K. (2017).
\newblock Calibrate fisheye lens using opencv.
\newblock \url{https://medium.com/@kennethjiang/calibrate-fisheye-lens-using-opencv-333b05afa0b0}.
\newblock Accessed: September 5, 2024
\bibAnnoteFile{jiang2017fisheye}

\bibitem[{Kang and Chen(2020)}]{kang2020fast}
Kang, H. and Chen, C. (2020).
\newblock Fast implementation of real-time fruit detection in apple orchards using deep learning.
\newblock \emph{Computers and Electronics in Agriculture} 168, 105108
\bibAnnoteFile{kang2020fast}

\bibitem[{Kingma and Ba(2014)}]{Adam2014}
Kingma, D.~P. and Ba, J. (2014).
\newblock Adam: A method for stochastic optimization.
\newblock \emph{3rd International Conference on Learning Representations, ICLR 2015 - Conference Track Proceedings}
\bibAnnoteFile{Adam2014}

\bibitem[{Krause et~al.(2023)Krause, Piepho, Dias, Singh, and Beavis}]{krause2023models}
Krause, M.~D., Piepho, H.-P., Dias, K.~O., Singh, A.~K., and Beavis, W.~D. (2023).
\newblock Models to estimate genetic gain of soybean seed yield from annual multi-environment field trials.
\newblock \emph{Theoretical and Applied Genetics} 136, 252
\bibAnnoteFile{krause2023models}

\bibitem[{Li et~al.(2024)Li, Magar, Chen, Lin, Wang, Yin et~al.}]{li2024soybeannet}
Li, J., Magar, R.~T., Chen, D., Lin, F., Wang, D., Yin, X., et~al. (2024).
\newblock Soybeannet: Transformer-based convolutional neural network for soybean pod counting from unmanned aerial vehicle (uav) images.
\newblock \emph{Computers and Electronics in Agriculture} 220, 108861
\bibAnnoteFile{li2024soybeannet}

\bibitem[{Li et~al.(2019)Li, Jia, Zhang, Khattak, Sun, Gao et~al.}]{li2019soybean}
Li, Y., Jia, J., Zhang, L., Khattak, A.~M., Sun, S., Gao, W., et~al. (2019).
\newblock Soybean seed counting based on pod image using two-column convolution neural network.
\newblock \emph{Ieee Access} 7, 64177--64185
\bibAnnoteFile{li2019soybean}

\bibitem[{Lu et~al.(2022)Lu, Du, Niu, Xing, Luo, Deng et~al.}]{lu2022soybean}
Lu, W., Du, R., Niu, P., Xing, G., Luo, H., Deng, Y., et~al. (2022).
\newblock Soybean yield preharvest prediction based on bean pods and leaves image recognition using deep learning neural network combined with grnn.
\newblock \emph{Frontiers in Plant Science} 12, 791256
\bibAnnoteFile{lu2022soybean}

\bibitem[{Maimaitijiang et~al.(2020)Maimaitijiang, Sagan, Sidike, Hartling, Esposito, and Fritschi}]{maimaitijiang2020soybean}
Maimaitijiang, M., Sagan, V., Sidike, P., Hartling, S., Esposito, F., and Fritschi, F.~B. (2020).
\newblock Soybean yield prediction from uav using multimodal data fusion and deep learning.
\newblock \emph{Remote Sensing of Environment} 237, 111599
\bibAnnoteFile{maimaitijiang2020soybean}

\bibitem[{Mathew et~al.(2023)Mathew, Delavarpour, Miranda, Stenger, Zhang, Aduteye et~al.}]{mathew2023novel}
Mathew, J., Delavarpour, N., Miranda, C., Stenger, J., Zhang, Z., Aduteye, J., et~al. (2023).
\newblock A novel approach to pod count estimation using a depth camera in support of soybean breeding applications.
\newblock \emph{Sensors} 23, 6506
\bibAnnoteFile{mathew2023novel}

\bibitem[{McGuire et~al.(2021)McGuire, Soman, Diers, and Chowdhary}]{mcguire2021}
McGuire, M., Soman, C., Diers, B., and Chowdhary, G. (2021).
\newblock High throughput soybean pod-counting with in-field robotic data collection and machine-vision based data analysis.
\newblock \emph{arXiv preprint arXiv:2105.10568}
\bibAnnoteFile{mcguire2021}

\bibitem[{Medic et~al.(2014)Medic, Atkinson, and Hurburgh}]{medic2014current}
Medic, J., Atkinson, C., and Hurburgh, C.~R. (2014).
\newblock Current knowledge in soybean composition.
\newblock \emph{Journal of the American Oil Chemists' Society} 91, 363--384
\bibAnnoteFile{medic2014current}

\bibitem[{Ning et~al.(2022)Ning, Zhao, and Zhang}]{ning2022extraction}
Ning, S., Zhao, Q., and Zhang, X. (2022).
\newblock Extraction of soybean pod features based on computer vision.
\newblock In \emph{International Conference on 5G for Future Wireless Networks} (Springer), 48--58
\bibAnnoteFile{ning2022extraction}

\bibitem[{Puhl et~al.(2021)Puhl, Bao, Sanz-Saez, and Chen}]{puhl2021infield}
Puhl, R.~B., Bao, Y., Sanz-Saez, A., and Chen, C. (2021).
\newblock Infield peanut pod counting using deep neural networks for yield estimation.
\newblock In \emph{2021 ASABE Annual International Virtual Meeting} (American Society of Agricultural and Biological Engineers), 1
\bibAnnoteFile{puhl2021infield}

\bibitem[{{Python Software Foundation}(2024)}]{python}
{Python Software Foundation} (2024).
\newblock \emph{Python Programming Language}.
\newblock Version 3.10.0
\bibAnnoteFile{python}

\bibitem[{Riera et~al.(2021)Riera, Carroll, Zhang, Shook, Ghosal, Gao et~al.}]{Riera2021}
Riera, L.~G., Carroll, M.~E., Zhang, Z., Shook, J.~M., Ghosal, S., Gao, T., et~al. (2021).
\newblock Deep multiview image fusion for soybean yield estimation in breeding applications.
\newblock \emph{Plant Phenomics} 2021, 1--12.
\newblock \doi{10.34133/2021/9846470}
\bibAnnoteFile{Riera2021}

\bibitem[{Sa et~al.(2016)Sa, Ge, Dayoub, Upcroft, Perez, and McCool}]{sa2016deepfruits}
Sa, I., Ge, Z., Dayoub, F., Upcroft, B., Perez, T., and McCool, C. (2016).
\newblock Deepfruits: A fruit detection system using deep neural networks.
\newblock \emph{Sensors} 16, 1222
\bibAnnoteFile{sa2016deepfruits}

\bibitem[{Sarkar et~al.(2024)Sarkar, Ganapathysubramanian, Singh, Fotouhi, Kar, Nagasubramanian et~al.}]{sarkar2024cyber}
Sarkar, S., Ganapathysubramanian, B., Singh, A., Fotouhi, F., Kar, S., Nagasubramanian, K., et~al. (2024).
\newblock Cyber-agricultural systems for crop breeding and sustainable production.
\newblock \emph{Trends in Plant Science} 29, 130--149
\bibAnnoteFile{sarkar2024cyber}

\bibitem[{Singh et~al.(2024)Singh, Balabaygloo, Bekee, Blair, Fey, Fotouhi et~al.}]{singh2024smart}
Singh, A.~K., Balabaygloo, B.~J., Bekee, B., Blair, S.~W., Fey, S., Fotouhi, F., et~al. (2024).
\newblock Smart connected farms and networked farmers to improve crop production, sustainability and profitability.
\newblock \emph{Frontiers in Agronomy}
\bibAnnoteFile{singh2024smart}

\bibitem[{Singh et~al.(2021{\natexlab{a}})Singh, Singh, and Singh}]{singh2021plant}
Singh, D.~P., Singh, A.~K., and Singh, A. (2021{\natexlab{a}}).
\newblock \emph{Plant Breeding and Cultivar Development} (Academic Press)
\bibAnnoteFile{singh2021plant}

\bibitem[{Singh et~al.(2021{\natexlab{b}})Singh, Singh, and Singh}]{singh2021plant_ch28}
Singh, D.~P., Singh, A.~K., and Singh, A. (2021{\natexlab{b}}).
\newblock \emph{Plant Breeding and Cultivar Development} (Academic Press).
\newblock Chapter 28
\bibAnnoteFile{singh2021plant_ch28}

\bibitem[{Singh et~al.(2021{\natexlab{c}})Singh, Singh, and Singh}]{singh2021plant_ch5}
Singh, D.~P., Singh, A.~K., and Singh, A. (2021{\natexlab{c}}).
\newblock \emph{Plant Breeding and Cultivar Development} (Academic Press).
\newblock Chapter 5
\bibAnnoteFile{singh2021plant_ch5}

\bibitem[{Singh et~al.(2021{\natexlab{d}})Singh, Singh, and Singh}]{singh2021plant_ch25}
Singh, D.~P., Singh, A.~K., and Singh, A. (2021{\natexlab{d}}).
\newblock \emph{Plant Breeding and Cultivar Development} (Academic Press).
\newblock Chapter 25
\bibAnnoteFile{singh2021plant_ch25}

\bibitem[{Song et~al.(2021)Song, Wang, Jiang, Wang, Tai, Wang et~al.}]{Song2021}
Song, Q., Wang, C., Jiang, Z., Wang, Y., Tai, Y., Wang, C., et~al. (2021).
\newblock Rethinking counting and localization in crowds: A purely point-based framework.
\newblock In \emph{2021 IEEE/CVF International Conference on Computer Vision (ICCV)} (IEEE), 3345--3354.
\newblock \doi{10.1109/ICCV48922.2021.00335}
\bibAnnoteFile{Song2021}

\bibitem[{Technow(2015)}]{technow2015r}
Technow, F. (2015).
\newblock R package mvnggrad: moving grid adjustment in plant breeding field trials.
\newblock \emph{R package version 0.1} 5
\bibAnnoteFile{technow2015r}

\bibitem[{Uzal et~al.(2018)Uzal, Grinblat, Nam{\'\i}as, Larese, Bianchi, Morandi et~al.}]{uzal2018seed}
Uzal, L.~C., Grinblat, G.~L., Nam{\'\i}as, R., Larese, M.~G., Bianchi, J.~S., Morandi, E.~N., et~al. (2018).
\newblock Seed-per-pod estimation for plant breeding using deep learning.
\newblock \emph{Computers and Electronics in Agriculture} 150, 196--204
\bibAnnoteFile{uzal2018seed}

\bibitem[{Wei and Molin(2020)}]{wei2020soybean}
Wei, M. C.~F. and Molin, J.~P. (2020).
\newblock Soybean yield estimation and its components: A linear regression approach.
\newblock \emph{Agriculture} 10, 348
\bibAnnoteFile{wei2020soybean}

\bibitem[{Xiang et~al.(2023)Xiang, Wang, Xu, Wang, and Liu}]{xiang2023yolo}
Xiang, S., Wang, S., Xu, M., Wang, W., and Liu, W. (2023).
\newblock Yolo pod: a fast and accurate multi-task model for dense soybean pod counting.
\newblock \emph{Plant Methods} 19, 8
\bibAnnoteFile{xiang2023yolo}

\bibitem[{Yang et~al.(2021)Yang, Zheng, He, Wu, Sun, and Wang}]{yang2021high}
Yang, S., Zheng, L., He, P., Wu, T., Sun, S., and Wang, M. (2021).
\newblock High-throughput soybean seeds phenotyping with convolutional neural networks and transfer learning.
\newblock \emph{Plant Methods} 17, 50
\bibAnnoteFile{yang2021high}

\bibitem[{Yang et~al.(2022)Yang, Zheng, Yang, Zhang, Wu, Sun et~al.}]{yang2022synthetic}
Yang, S., Zheng, L., Yang, H., Zhang, M., Wu, T., Sun, S., et~al. (2022).
\newblock A synthetic datasets based instance segmentation network for high-throughput soybean pods phenotype investigation.
\newblock \emph{Expert Systems with Applications} 192, 116403
\bibAnnoteFile{yang2022synthetic}

\bibitem[{Young et~al.(2024)Young, Chiranjeevi, Elango, Sarkar, Singh, Singh et~al.}]{young2024soybean}
Young, T.~J., Chiranjeevi, S., Elango, D., Sarkar, S., Singh, A.~K., Singh, A., et~al. (2024).
\newblock Soybean canopy stress classification using 3d point cloud data.
\newblock \emph{Agronomy} 14, 1181
\bibAnnoteFile{young2024soybean}

\bibitem[{Young et~al.(2023)Young, Jubery, Carley, Carroll, Sarkar, Singh et~al.}]{young2023canopy}
Young, T.~J., Jubery, T.~Z., Carley, C.~N., Carroll, M., Sarkar, S., Singh, A.~K., et~al. (2023).
\newblock “canopy fingerprints” for characterizing three-dimensional point cloud data of soybean canopies.
\newblock \emph{Frontiers in plant science} 14, 1141153
\bibAnnoteFile{young2023canopy}

\bibitem[{Yu et~al.(2016)Yu, Li, Schmitz, Tian, Greenberg, and Diers}]{yu2016development}
Yu, N., Li, L., Schmitz, N., Tian, L.~F., Greenberg, J.~A., and Diers, B.~W. (2016).
\newblock Development of methods to improve soybean yield estimation and predict plant maturity with an unmanned aerial vehicle based platform.
\newblock \emph{Remote Sensing of Environment} 187, 91--101
\bibAnnoteFile{yu2016development}

\bibitem[{Yu et~al.(2024)Yu, Wang, Ye, Liufu, Lu, Zhu et~al.}]{yu2024accurate}
Yu, Z., Wang, Y., Ye, J., Liufu, S., Lu, D., Zhu, X., et~al. (2024).
\newblock Accurate and fast implementation of soybean pod counting and localization from high-resolution image.
\newblock \emph{Frontiers in Plant Science} 15, 1320109
\bibAnnoteFile{yu2024accurate}

\bibitem[{Zhang et~al.(2023)Zhang, Lu, Ma, Hu, Zhang, Ning et~al.}]{zhang2023high}
Zhang, C., Lu, X., Ma, H., Hu, Y., Zhang, S., Ning, X., et~al. (2023).
\newblock High-throughput classification and counting of vegetable soybean pods based on deep learning.
\newblock \emph{Agronomy} 13, 1154
\bibAnnoteFile{zhang2023high}

\bibitem[{Zhao et~al.(2023)Zhao, Kaga, Yamada, Komatsu, Hirata, Kikuchi et~al.}]{Zhao2023}
Zhao, J., Kaga, A., Yamada, T., Komatsu, K., Hirata, K., Kikuchi, A., et~al. (2023).
\newblock Improved field-based soybean seed counting and localization with feature level considered.
\newblock \emph{Plant Phenomics} 5.
\newblock \doi{10.34133/plantphenomics.0026}
\bibAnnoteFile{Zhao2023}

\end{thebibliography}


\section*{Tables and Figure captions}


\begin{subfigure}
\setcounter{figure}{3}
\setcounter{subfigure}{0}
    \centering
    \begin{minipage}[b]{0.5\textwidth}
        \includegraphics[width=\linewidth]{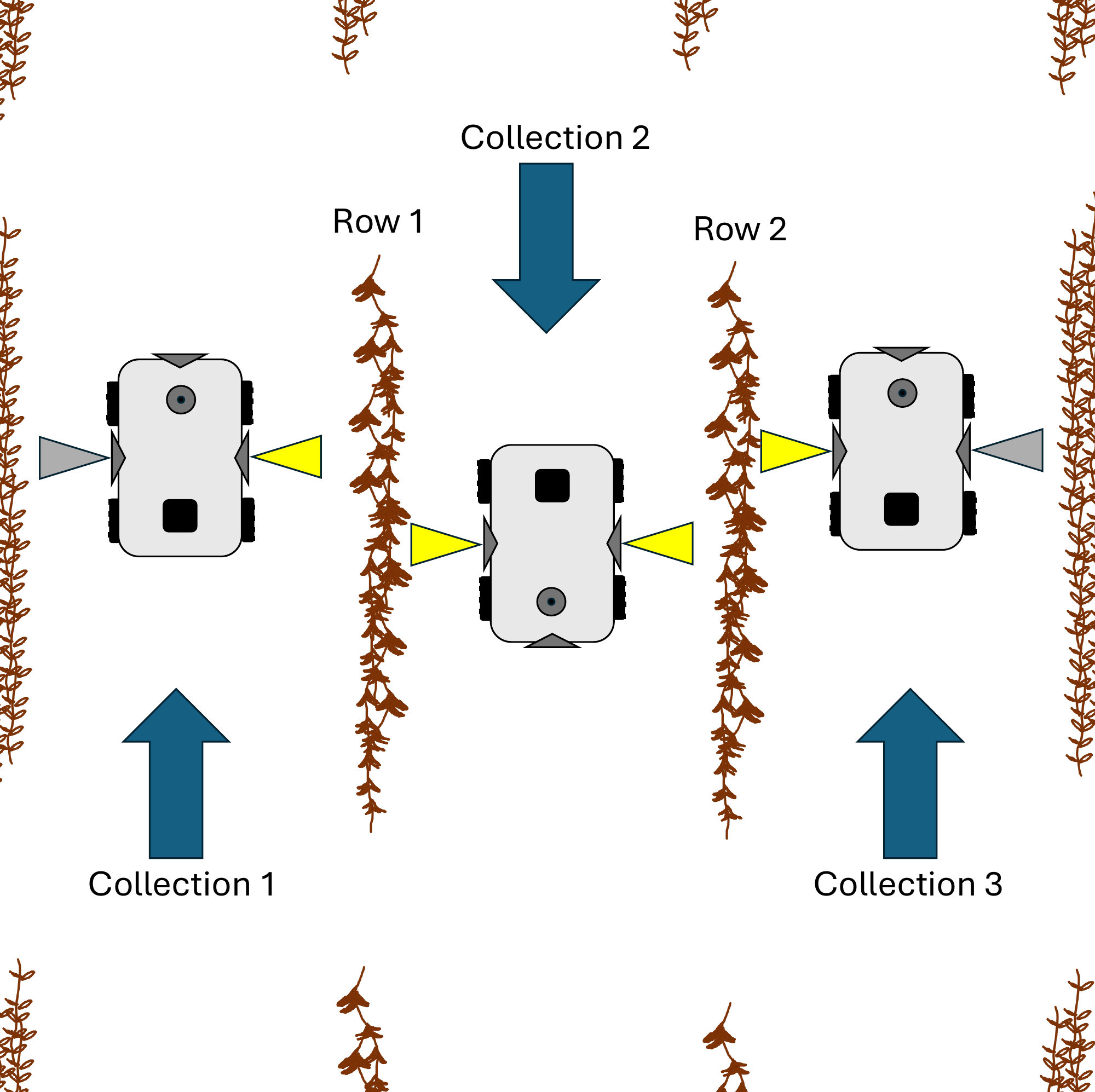}
        \caption{Data collection for a single plot.}
        \label{fig:plotCollection}
    \end{minipage}  
\setcounter{figure}{3}
\setcounter{subfigure}{1}
    \centering
    \begin{minipage}[b]{0.9\textwidth}
        \includegraphics[width=\linewidth]{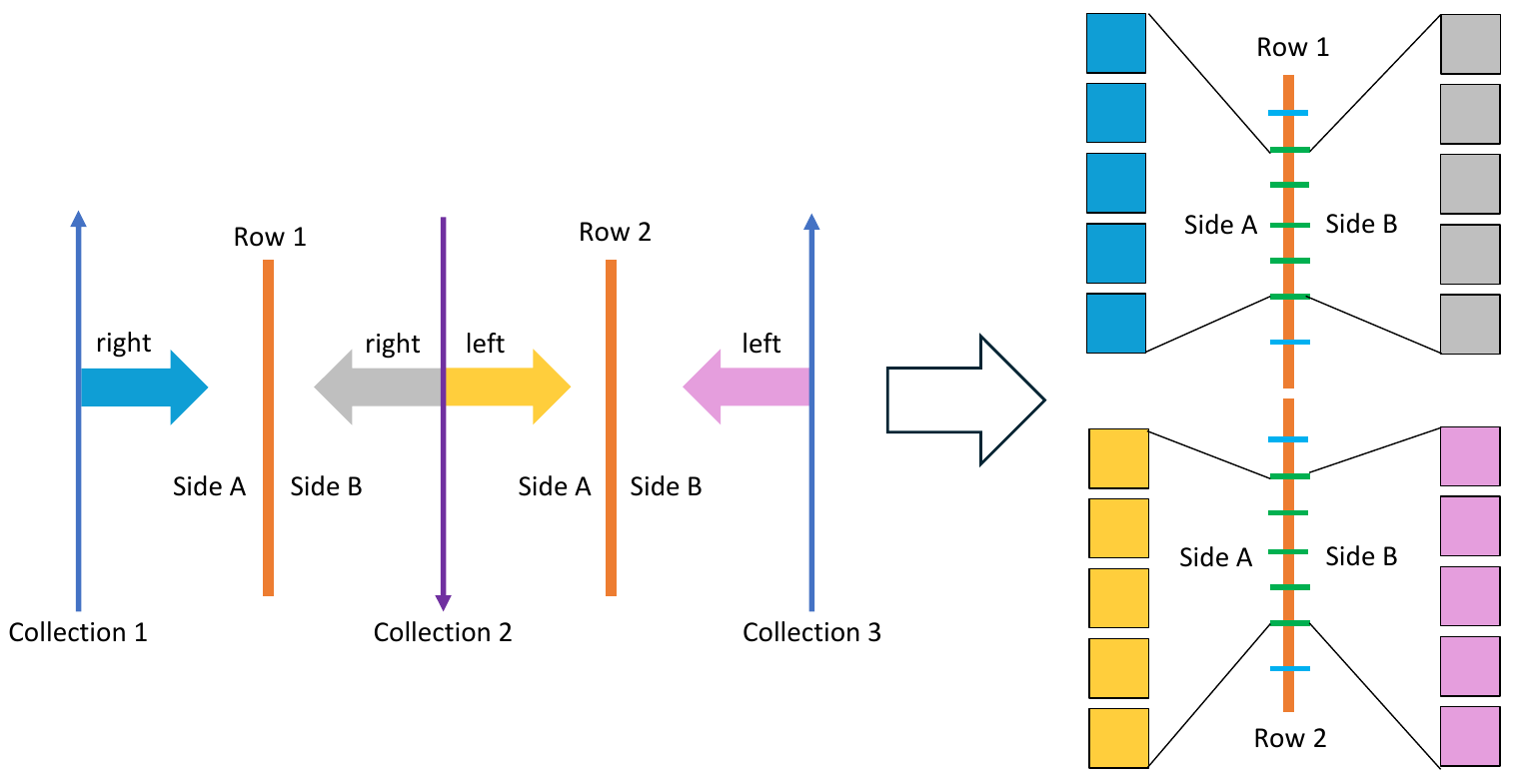}
        \caption{Creation of splitters for a plot.}
        \label{fig:sampling}
    \end{minipage}
\setcounter{figure}{3}
\setcounter{subfigure}{-1}
    \caption{ \textbf{(A)} demonstrates the data collection process for a single plot. The yellow-highlighted cameras represent video sections belonging to the center plot. Grey-highlighted cameras represent video of other plots. Three collections in total are needed to fully capture a single plot. Post-processing with Python organizes these videos into their respective plots. Arrows represent the direction of robot movement. \textbf{(B)} Image data sampling process for a single plot. Each row was equidistantly divided into eight sections using seven splitters, with images from the middle five splitters chosen for analysis. (The first and last splitters were excluded.) The two rows of the same plot were connected and treated as one single row, resulting in ten images per side and twenty images per plot. }
    \label{fig:terrasentia_robot_plotCollection}
\end{subfigure}

\begin{figure}[h!tb]
    \centering
    \includegraphics[width=0.8\textwidth]{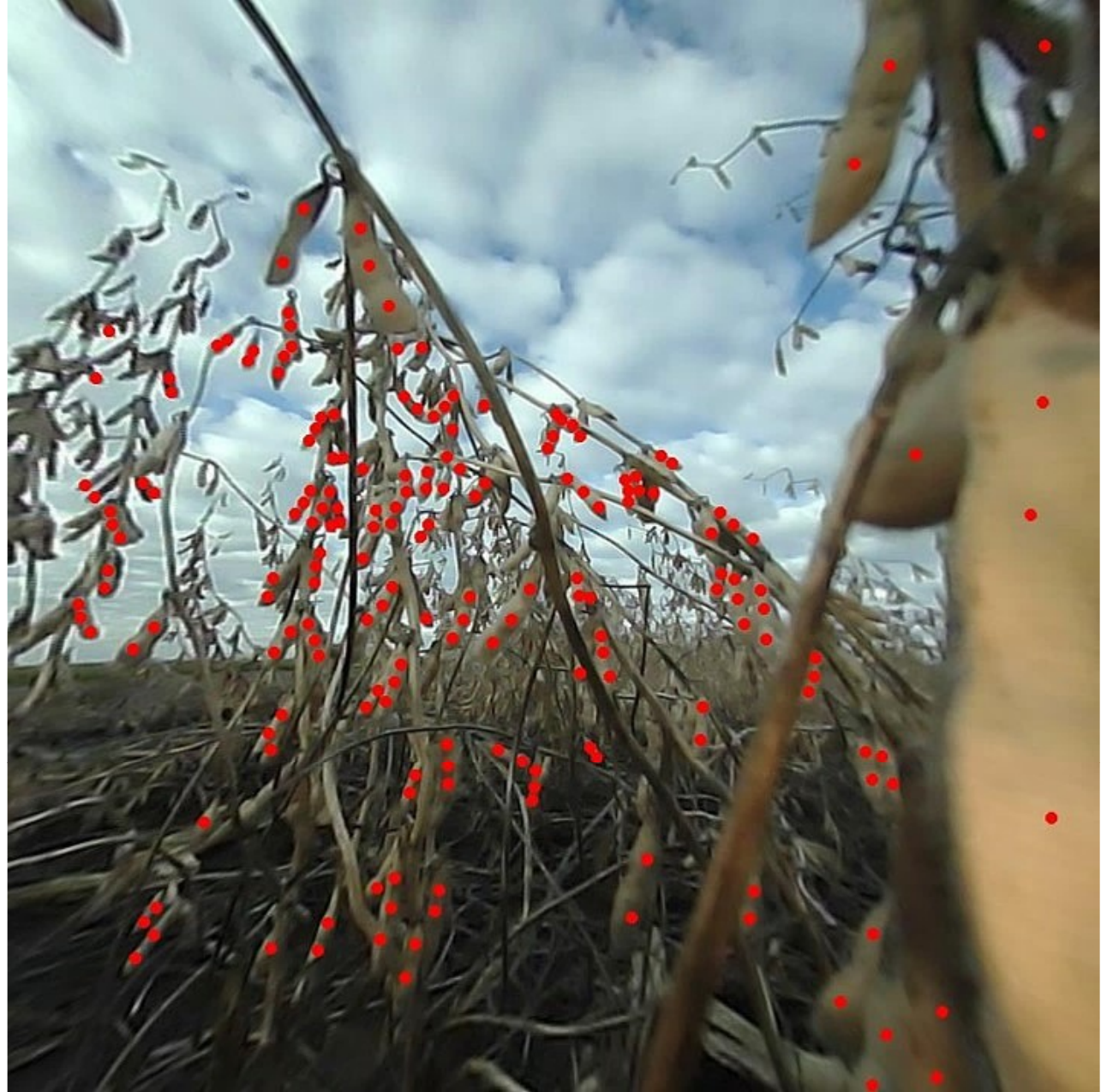}
    \caption{Example of an expertly annotated image. All seeds clearly discernible to the naked eye were annotated using point annotations. This image represents a frame that has been calibrated for fisheye distortion and has been cropped.}
    \label{fig:seed_annotations}
\end{figure}

\begin{figure}[h!tb]
    \centering
    \includegraphics[width=0.95\textwidth]{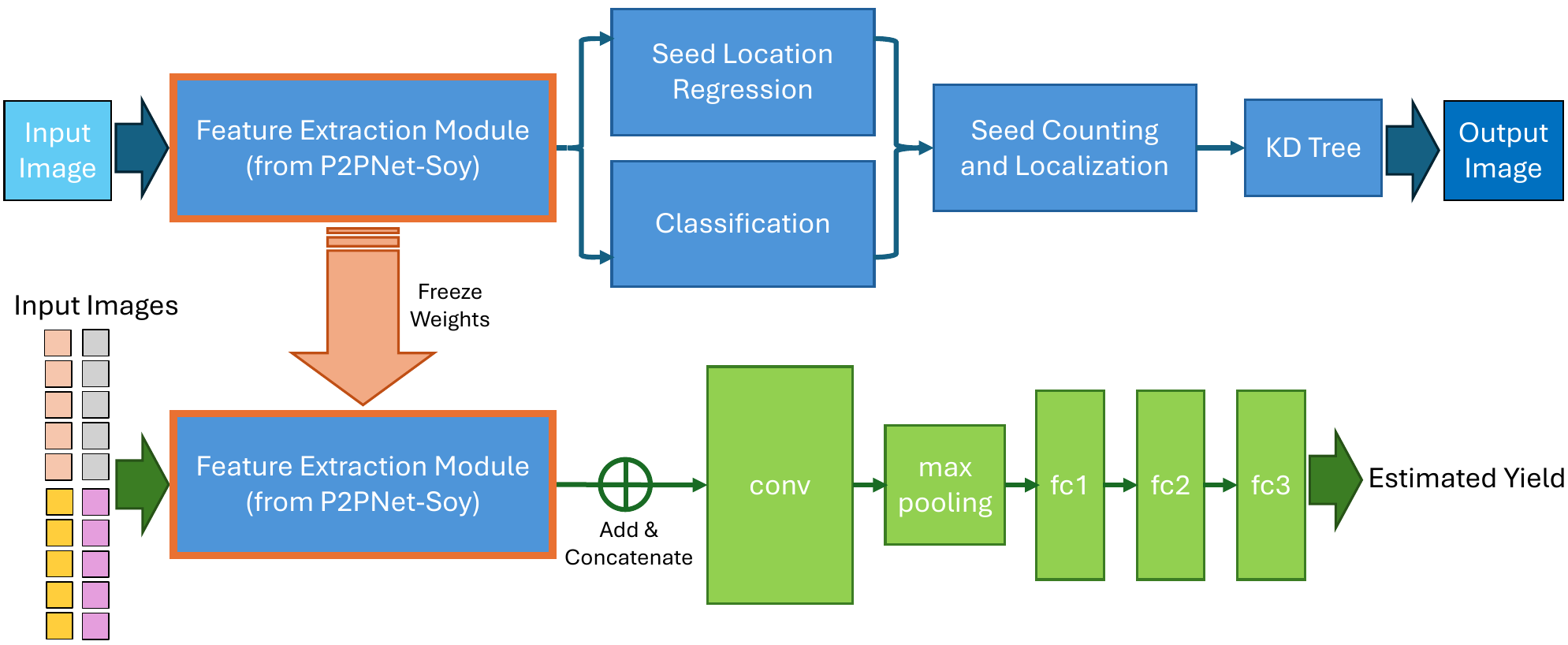}
    \caption{Our architecture, P2PNet-Yield, for soybean yield estimation. Training consists of two phases: first, training the P2PNet-Soy model so that its backbone (used as our Feature Extraction Module) can extract useful information related to soybean seeds in the foreground; second, training our Yield Regression Module to estimate yield values from the output feature maps of the Feature Extraction Module.}
    \label{fig:yield_architecture}
\end{figure}

\begin{subfigure}
\setcounter{figure}{6}
\setcounter{subfigure}{0}
    \centering
    \begin{minipage}[b]{0.45\textwidth}
        \includegraphics[width=\linewidth]{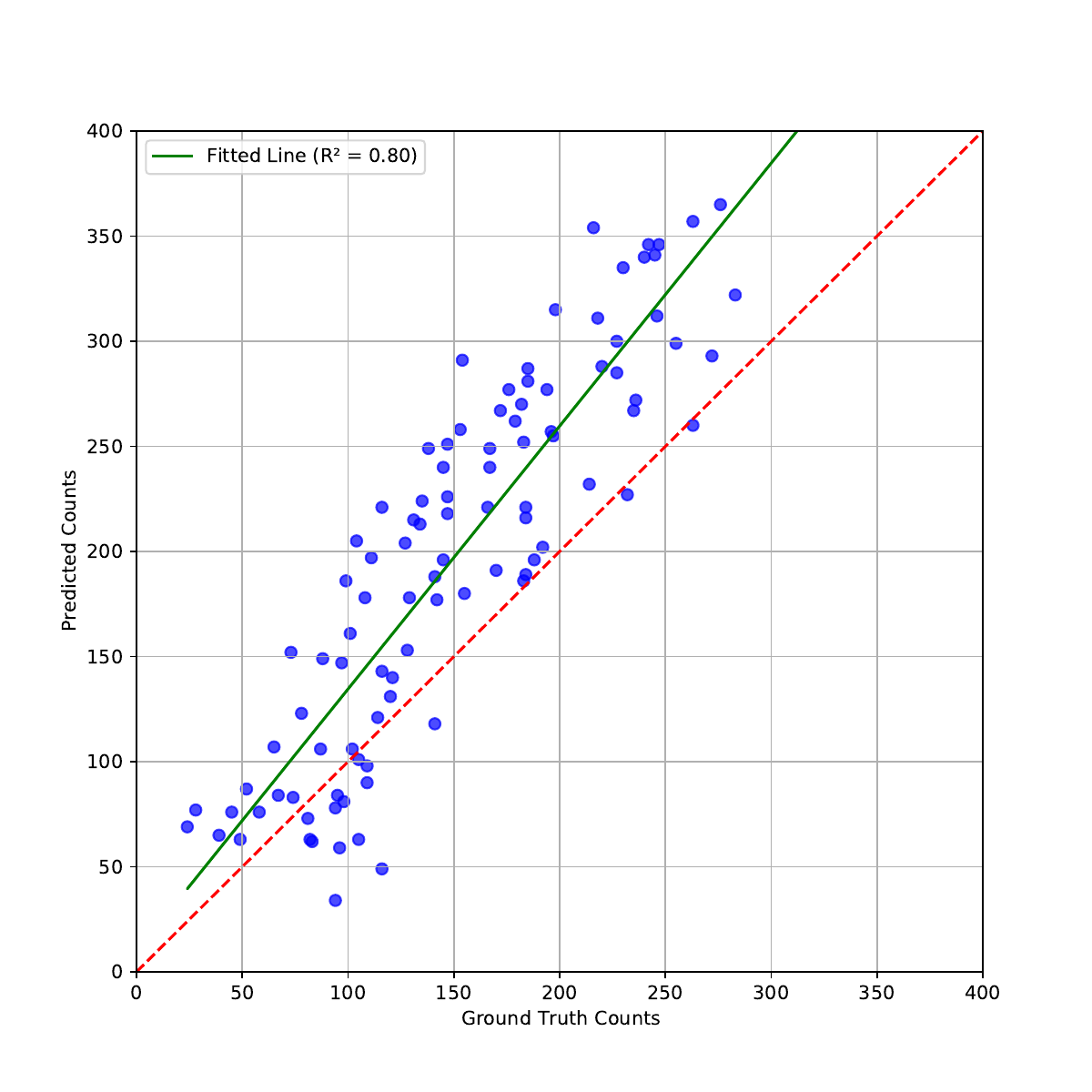}
        \caption{ISU\_NO\_AUG}
        \label{fig:sc_correlation_GTvsP_1}
    \end{minipage}  
\setcounter{figure}{6}
\setcounter{subfigure}{1}
    \centering
    \begin{minipage}[b]{0.45\textwidth}
        \includegraphics[width=\linewidth]{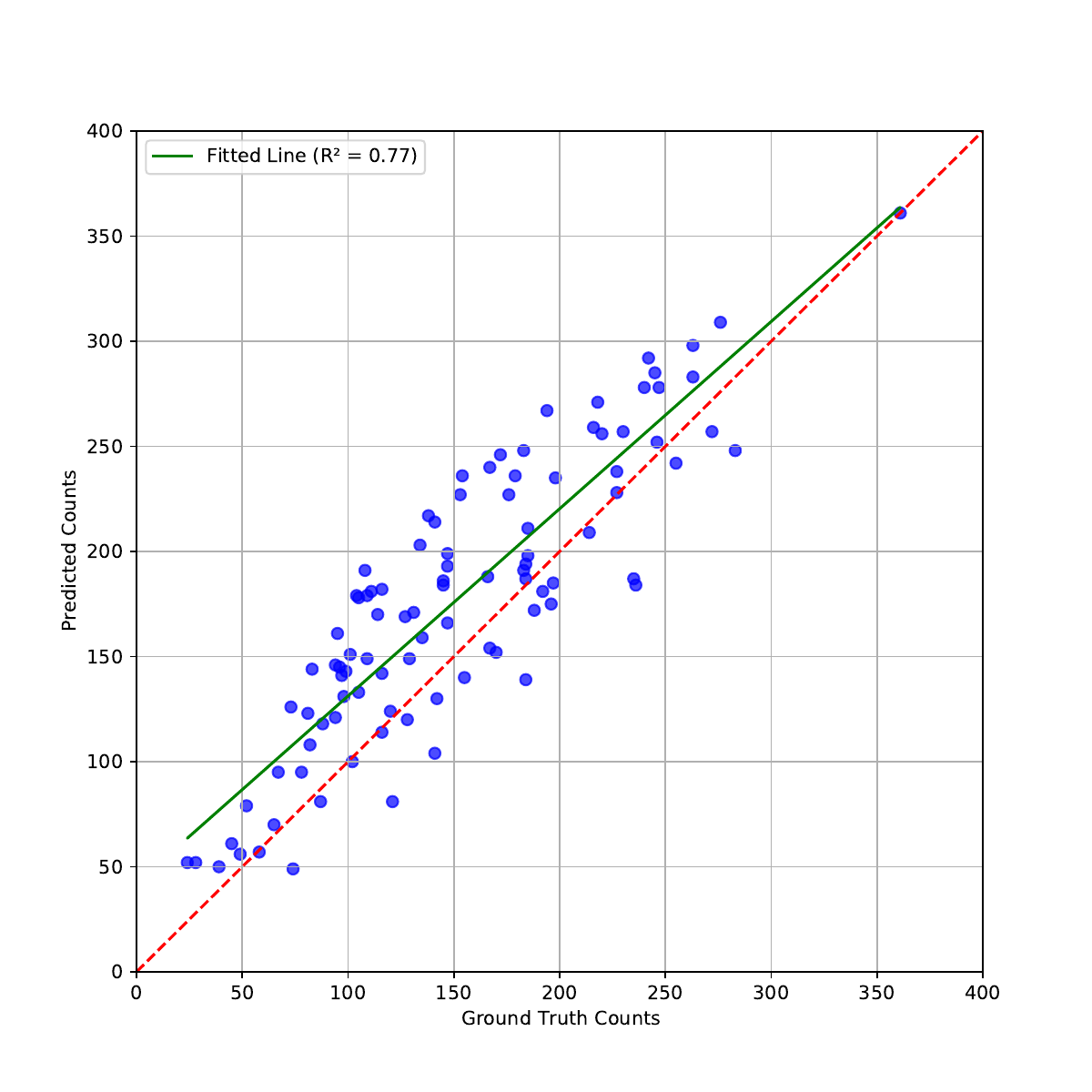}
        \caption{MIX\_NO\_AUG}
        \label{fig:sc_correlation_GTvsP_2}
    \end{minipage}
\setcounter{figure}{6}
\setcounter{subfigure}{2}
    \centering
    \begin{minipage}[b]{0.45\textwidth}
        \includegraphics[width=\linewidth]{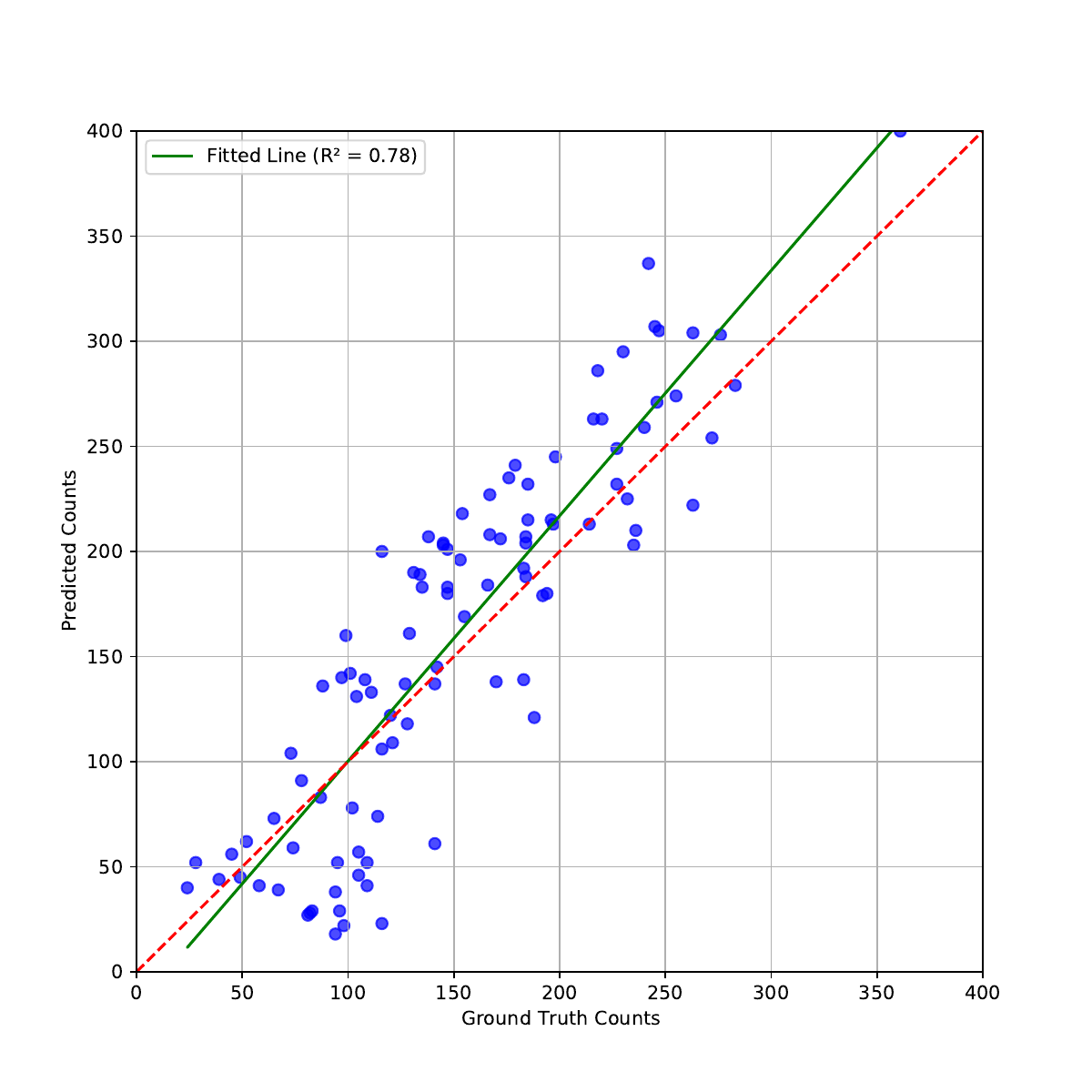}
        \caption{ISU\_AUG}
        \label{fig:sc_correlation_GTvsP_3}
    \end{minipage}  
\setcounter{figure}{6}
\setcounter{subfigure}{3}
    \centering
    \begin{minipage}[b]{0.45\textwidth}
        \includegraphics[width=\linewidth]{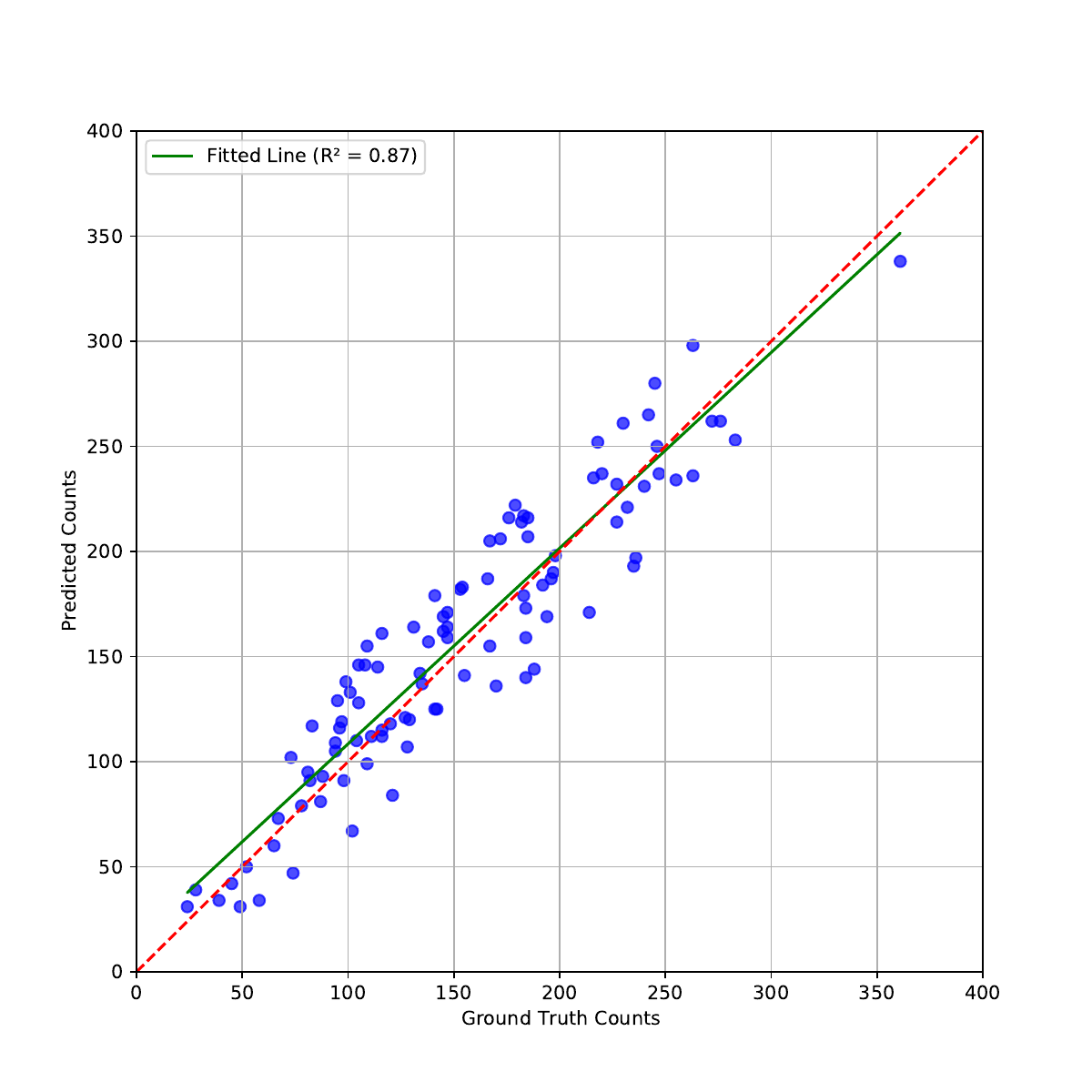}
        \caption{MIX\_AUG}
        \label{fig:sc_correlation_GTvsP_4}
    \end{minipage}  
\setcounter{figure}{6}
\setcounter{subfigure}{-1}
    \caption{Correlations between ground truth and estimated seed counts of models trained on different combinations of the datasets. The combination details can be found in Section~\ref{subsec:seed_counting}. Results show that the model trained on mixed datasets with data augmentation performs the best.}
    \label{fig:sc_correlation_GTvsP}
\end{subfigure}

\begin{table}[h!tb]
\centering
\caption{Testing results (MSE, MAE and MAPE) of models trained on different combinations of the datasets.}
\label{tab:SC_test_accuracy}
\begin{tabular}{cccc}
\hline
\textbf{Combination}  & \textbf{MSE} & \textbf{MAE} & \textbf{MAPE (\%)} \\ \hline
\textbf{ISU\_NO\_AUG} & 4277.83      & 54.93        & 41.26              \\ \hline
\textbf{MIX\_NO\_AUG} & 1734.11      & 34.89        & 28.72              \\ \hline
\textbf{ISU\_AUG}     & 1858.16      & 36.26        & 28.64              \\ \hline
\textbf{MIX\_AUG}     & 596.94       & 20.54        & 15.50              \\ \hline
\end{tabular}
\end{table}

\begin{subfigure}
\setcounter{figure}{7}
\setcounter{subfigure}{0}
    \centering
    \begin{minipage}[b]{0.45\textwidth}
        \includegraphics[width=\linewidth]{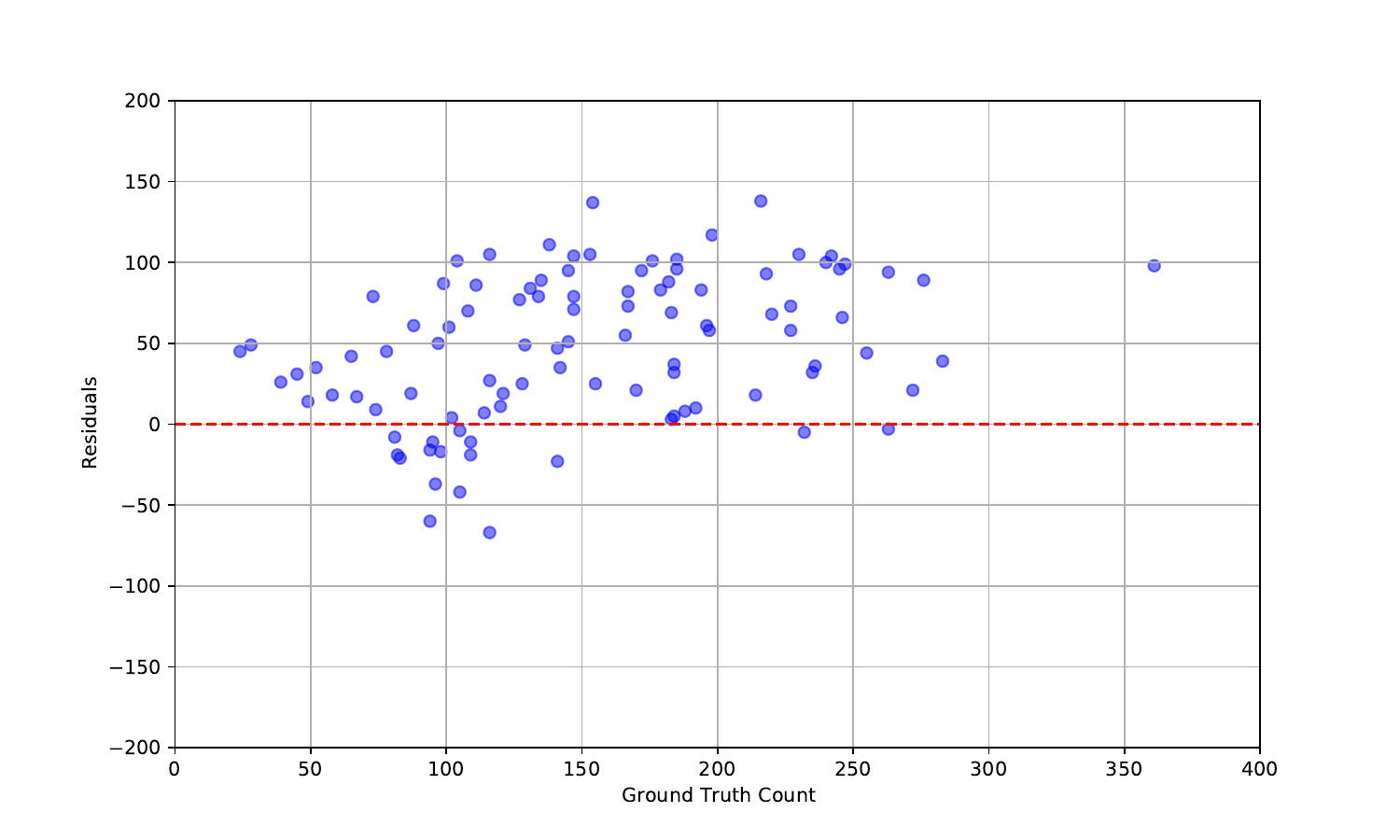}
        \caption{ISU\_NO\_AUG}
        \label{fig:sc_residual_1}
    \end{minipage}  
\setcounter{figure}{7}
\setcounter{subfigure}{1}
    \centering
    \begin{minipage}[b]{0.45\textwidth}
        \includegraphics[width=\linewidth]{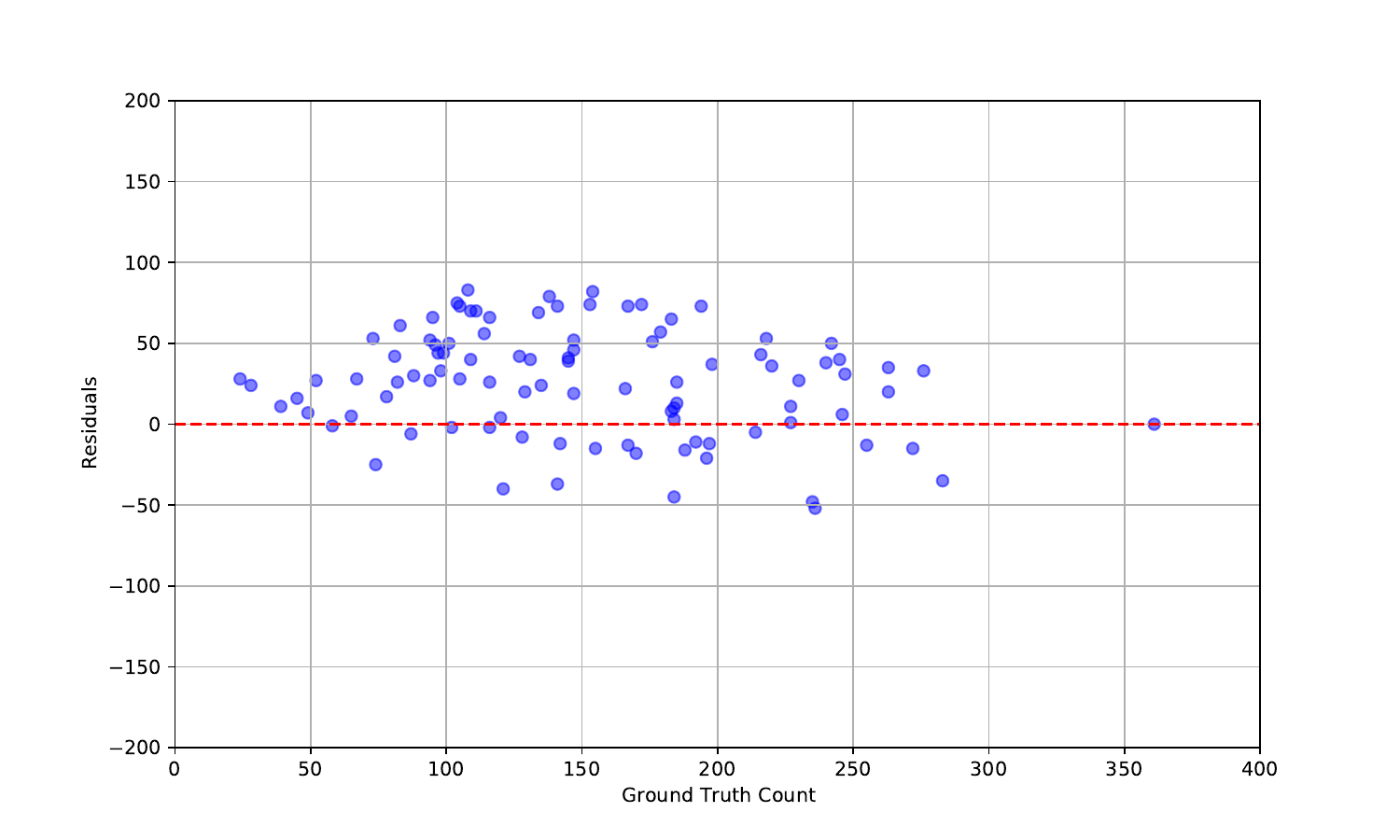}
        \caption{MIX\_NO\_AUG}
        \label{fig:sc_residual_2}
    \end{minipage}
\setcounter{figure}{7}
\setcounter{subfigure}{2}
    \centering
    \begin{minipage}[b]{0.45\textwidth}
        \includegraphics[width=\linewidth]{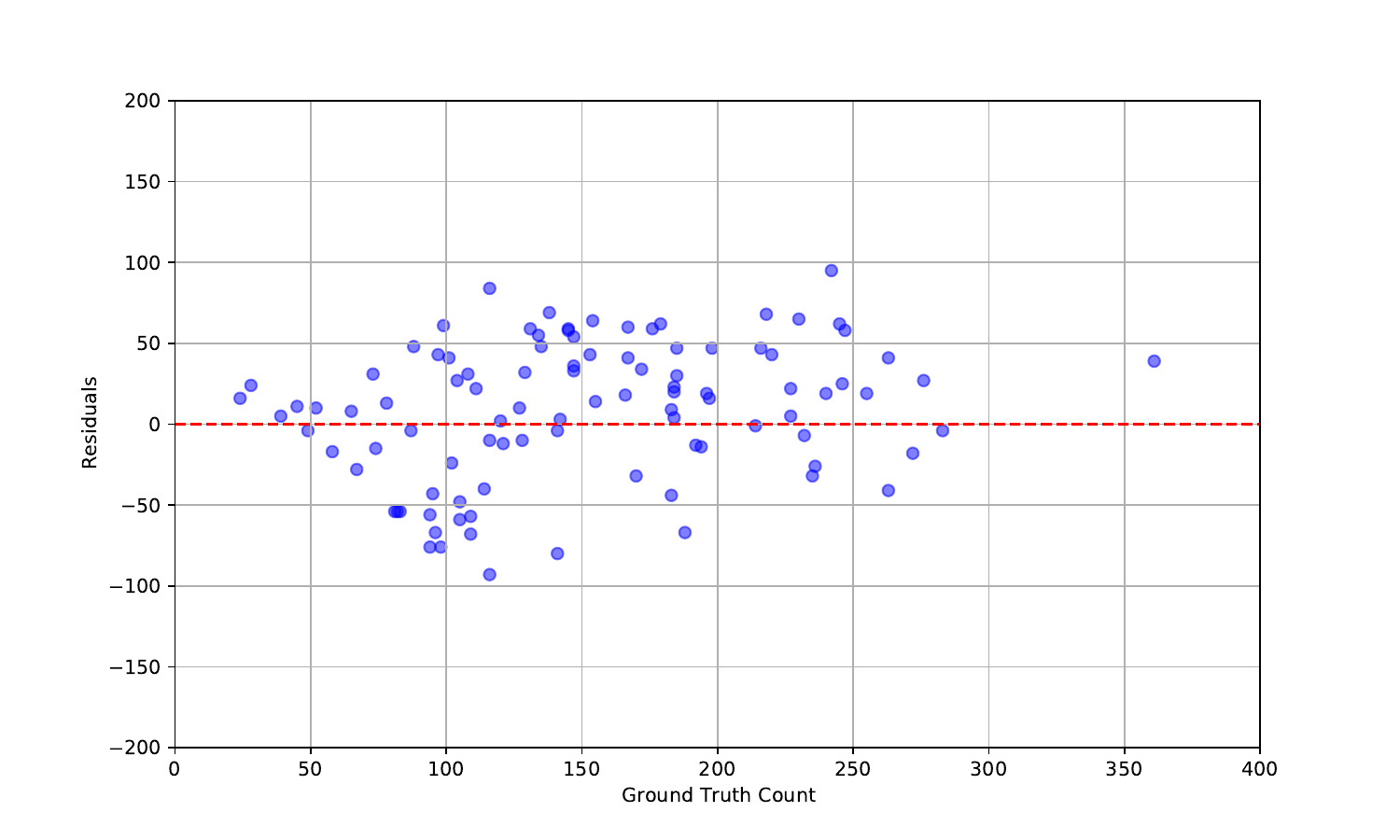}
        \caption{ISU\_AUG}
        \label{fig:sc_residual_3}
    \end{minipage}  
\setcounter{figure}{7}
\setcounter{subfigure}{3}
    \centering
    \begin{minipage}[b]{0.45\textwidth}
        \includegraphics[width=\linewidth]{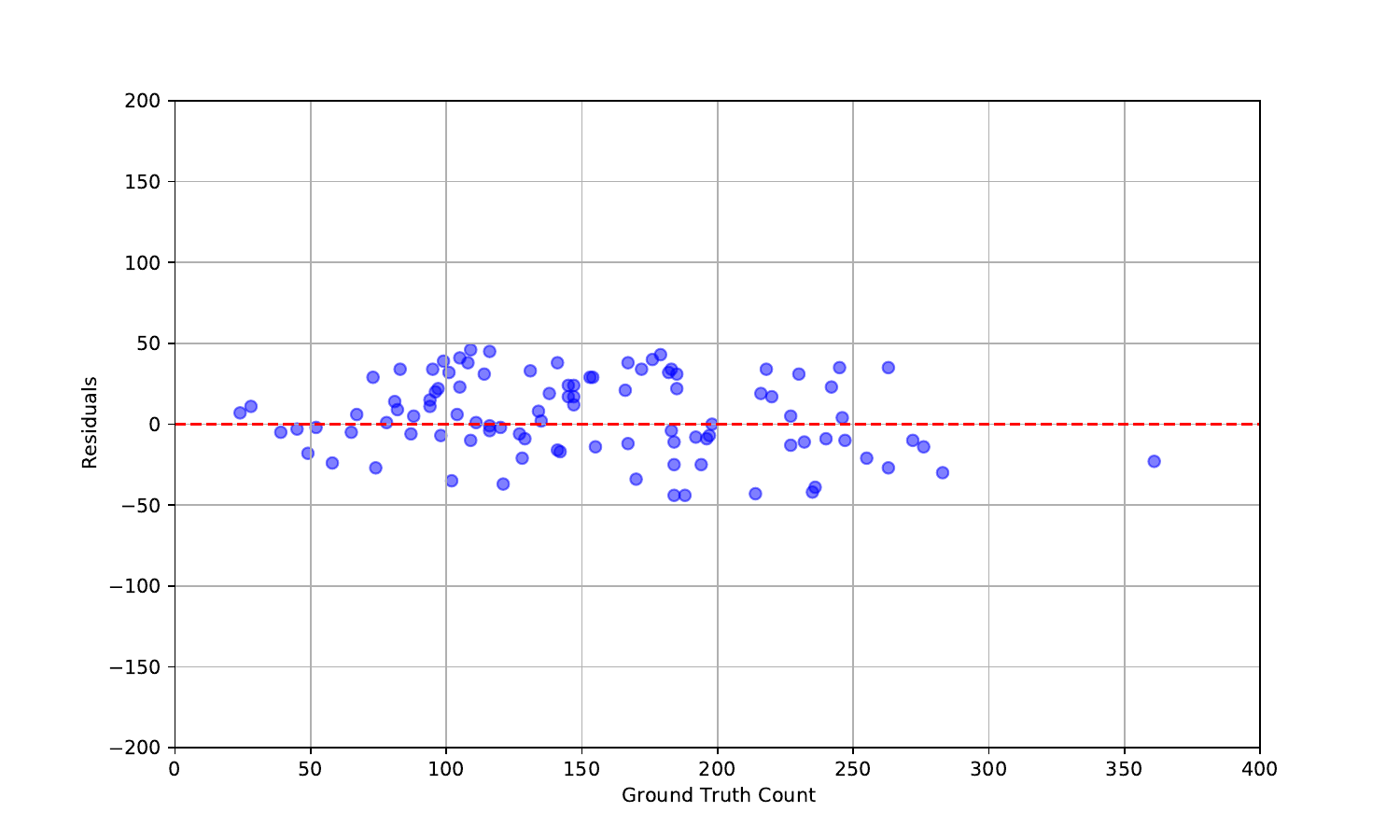}
        \caption{MIX\_AUG}
        \label{fig:sc_residual_4}
    \end{minipage}  
\setcounter{figure}{7}
\setcounter{subfigure}{-1}
    \caption{Residual plots of models trained on different combinations of the datasets. The combination details can be found in Section~\ref{subsec:seed_counting}. Results show that the model trained on mixed datasets with data augmentation performs the best.}
    \label{fig:sc_residual}
\end{subfigure}

\begin{figure}[h!tb]
    \centering
    \includegraphics[width=0.5\textwidth]{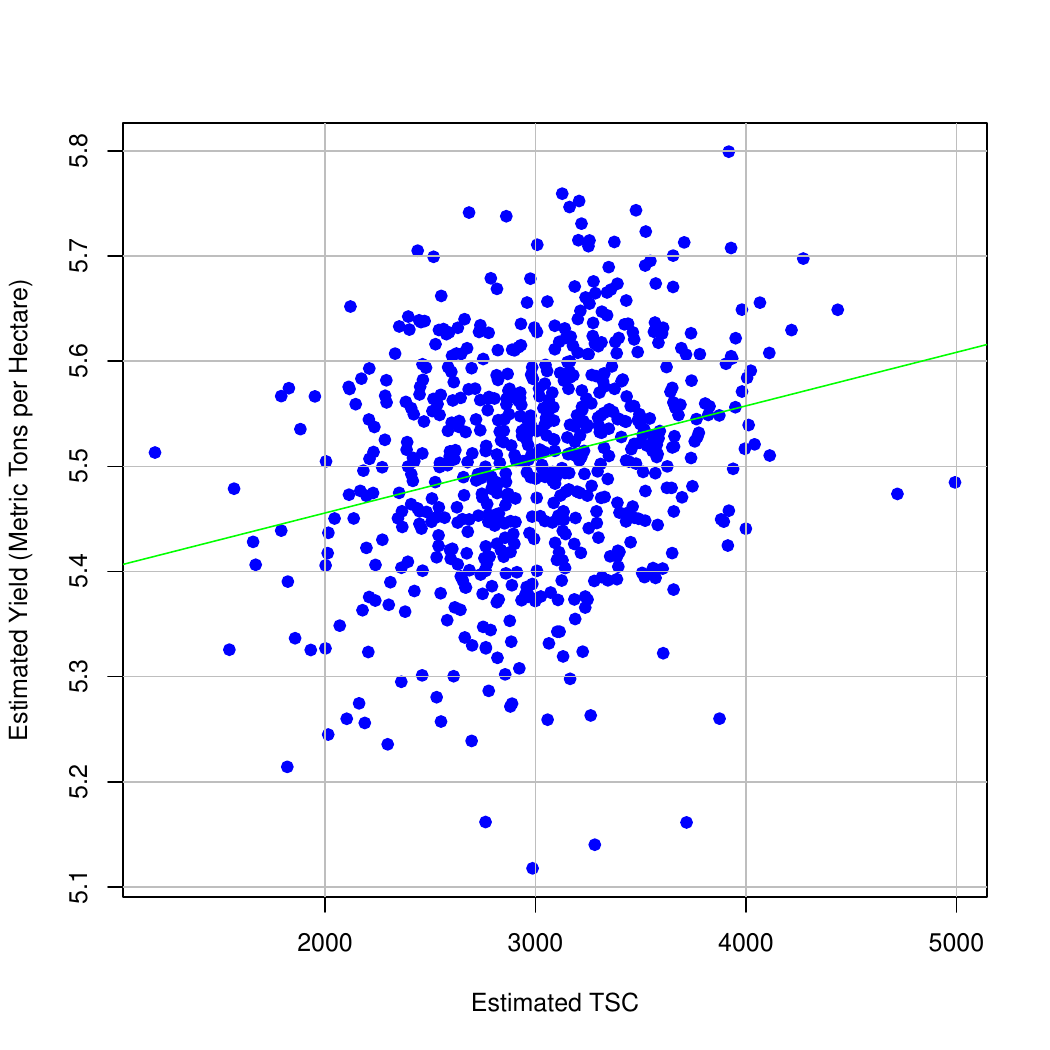}
    \caption{Correlation between estimated TSC and estimated yield for the 650 plots from the 2023 F7 field. $R^2$ value of 0.06 and a correlation coefficient of 0.25.}
    \label{fig:yield_vs_TSC}
\end{figure}

\begin{subfigure}
\setcounter{figure}{9}
\setcounter{subfigure}{0}
    \centering
    \begin{minipage}[b]{0.6\textwidth}
        \includegraphics[width=\linewidth]{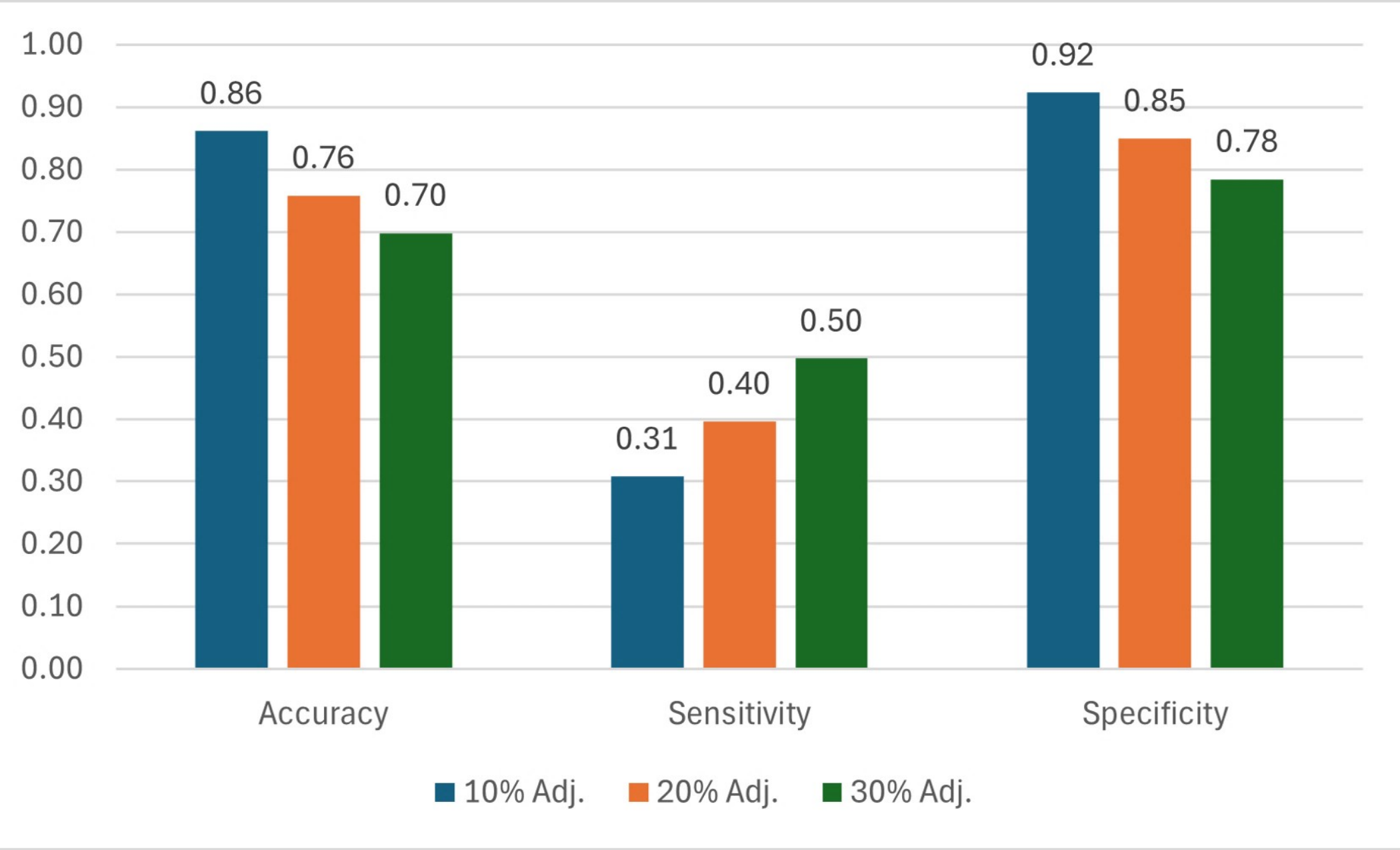}
        \caption{Ranking scores using TSC.}
        \label{fig:acc_sens_spec_1}
    \end{minipage}  
\setcounter{figure}{9}
\setcounter{subfigure}{1}
    \centering
    \begin{minipage}[b]{0.6\textwidth}
        \includegraphics[width=\linewidth]{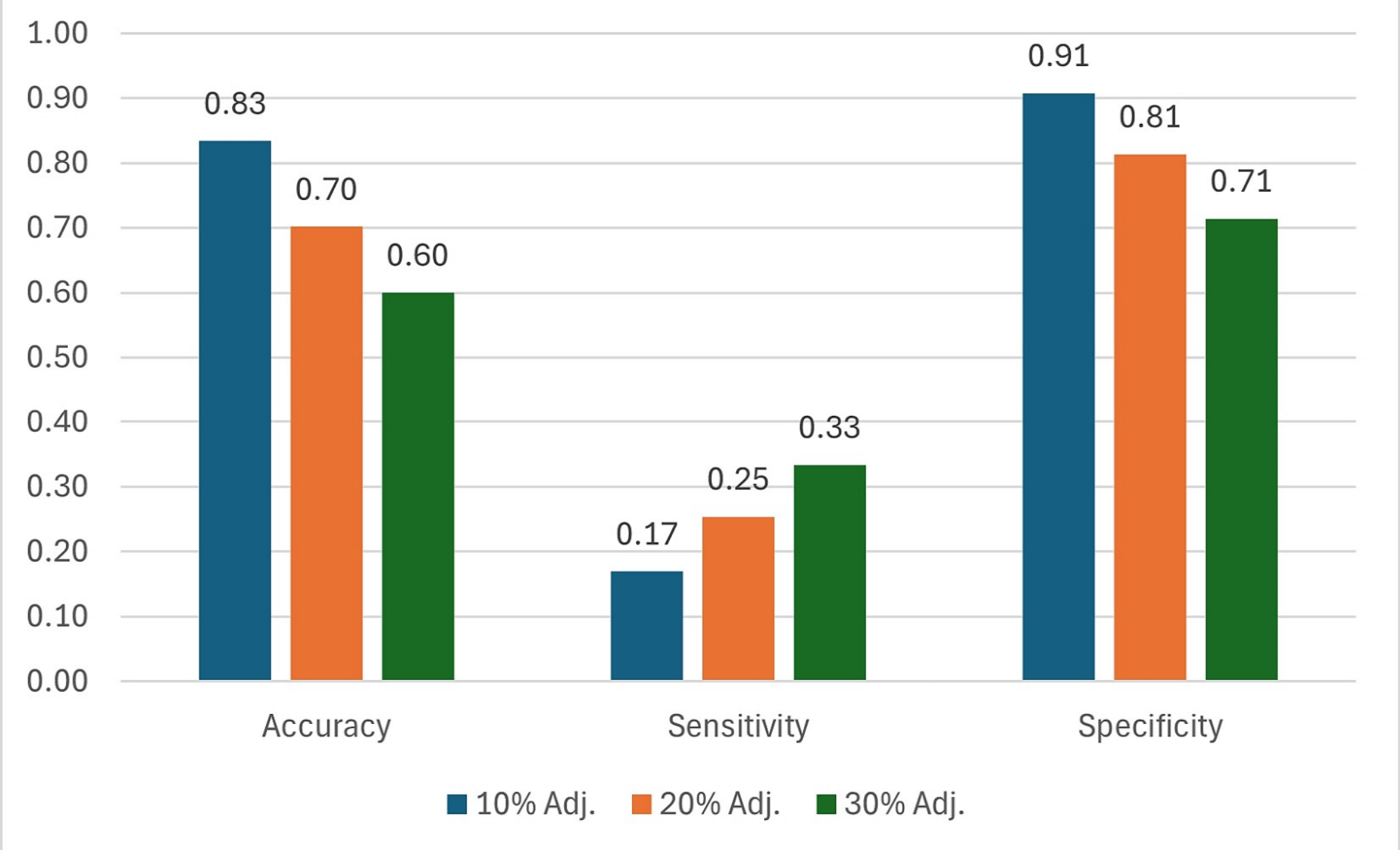}
        \caption{Ranking scores using estimated yield.}
        \label{fig:acc_sens_spec_2}
    \end{minipage}
\setcounter{figure}{9}
\setcounter{subfigure}{-1}
    \caption{Ranking scores for 10\%, 20\%, and 30\% selection thresholds using \textbf{(A)} TSC and \textbf{(B)} estimated yield. Scores represent spatially adjusted ground truth yields, TSC and estimated yields.}
    \label{fig:acc_sens_spec}
\end{subfigure}

\begin{table}[h!tb]
\centering
\caption{True positive (TP), true negative (TN), false positive (FP), and false negative (FN) values were calculated using a 10\%, 20\%, and 30\% selection threshold based on estimated TSC and estimated yield ranking.}
\label{tab:tp_tn_fp_fn}
\begin{tabular}{lccccc}
\hline
\textbf &  & \textbf{10\% Threshold}  &  \textbf{20\% Threshold} & \textbf{30\% Threshold} \\ \hline
\multirow{4}{*}{\textbf{Estimated TSC}} & \textbf{TP} & 20 & 52 & 97 \\ 
                                   & \textbf{TN} & 540 & 441 & 357 \\ 
                                   & \textbf{FP} & 45 & 78 & 98 \\ 
                                   & \textbf{FN} & 45 & 79 & 98 \\ \hline
\multirow{4}{*}{\textbf{Estimated Yield}} & \textbf{TP} & 11 & 33 & 65 \\ 
                                   & \textbf{TN} & 531 & 423 & 325 \\ 
                                   & \textbf{FP} & 54 & 97 & 130 \\ 
                                   & \textbf{FN} & 54 & 97 & 130 \\ \hline
\end{tabular}
\end{table}

\begin{subfigure}
\setcounter{figure}{10}
\setcounter{subfigure}{0}
    \centering
    \begin{minipage}[b]{0.5\textwidth}
        \includegraphics[width=\linewidth]{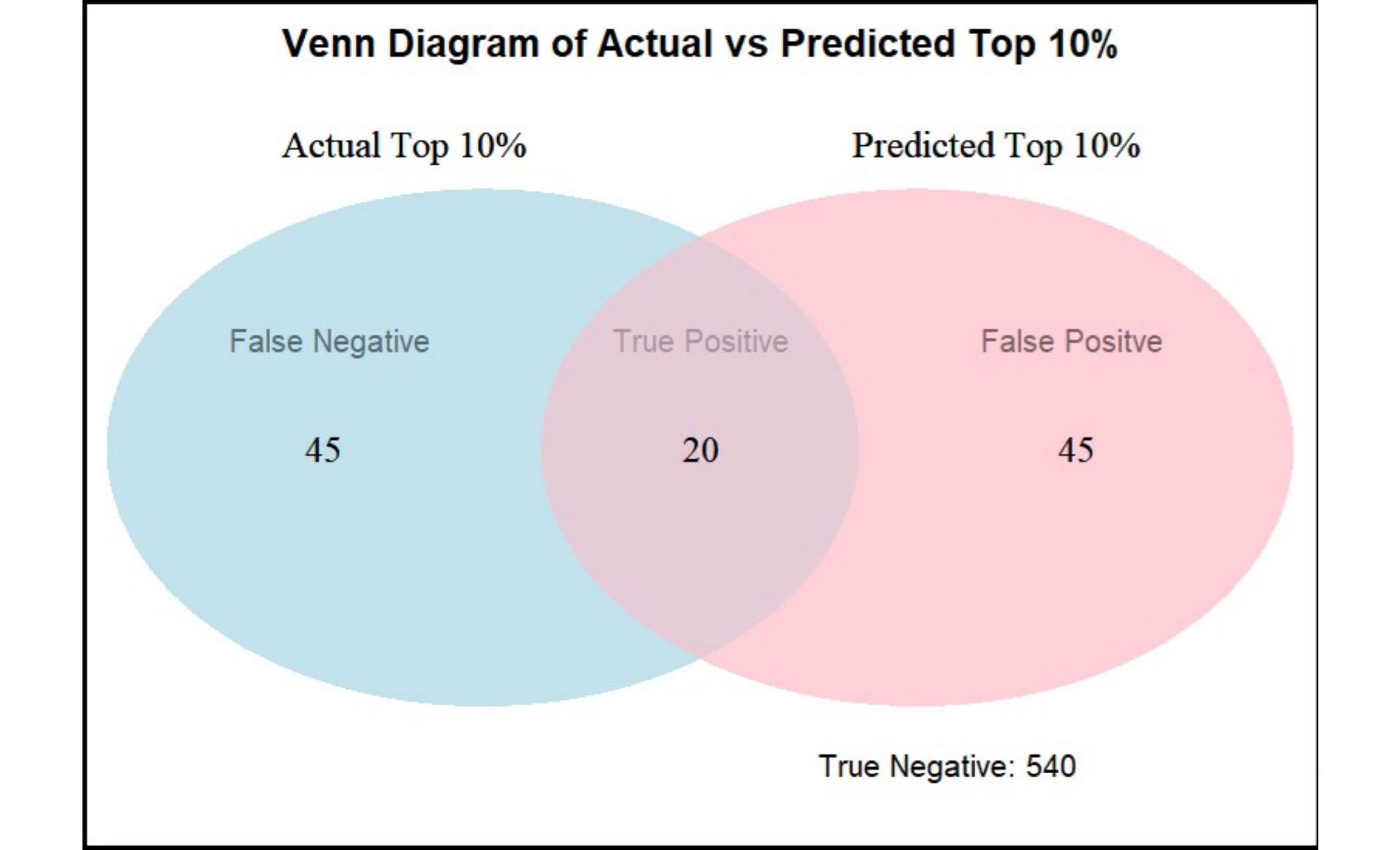}
        \caption{10\% selection threshold}
        \label{fig:venn diagram_1}
    \end{minipage}  
\setcounter{figure}{10}
\setcounter{subfigure}{1}
    \centering
    \begin{minipage}[b]{0.5\textwidth}
        \includegraphics[width=\linewidth]{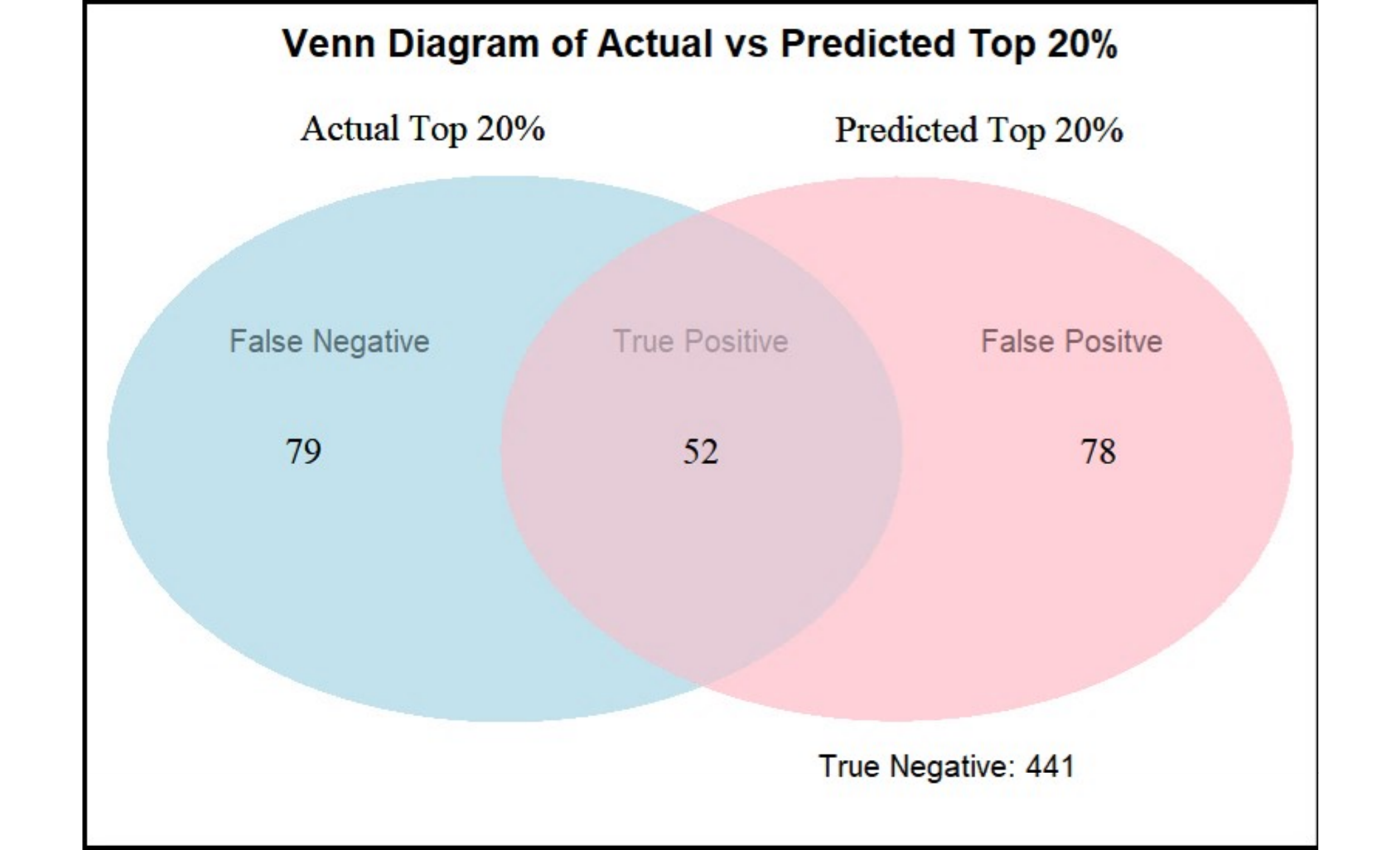}
        \caption{20\% selection threshold}
        \label{fig:venn diagram_2}
    \end{minipage}
\setcounter{figure}{10}
\setcounter{subfigure}{2}
    \centering
    \begin{minipage}[b]{0.5\textwidth}
        \includegraphics[width=\linewidth]{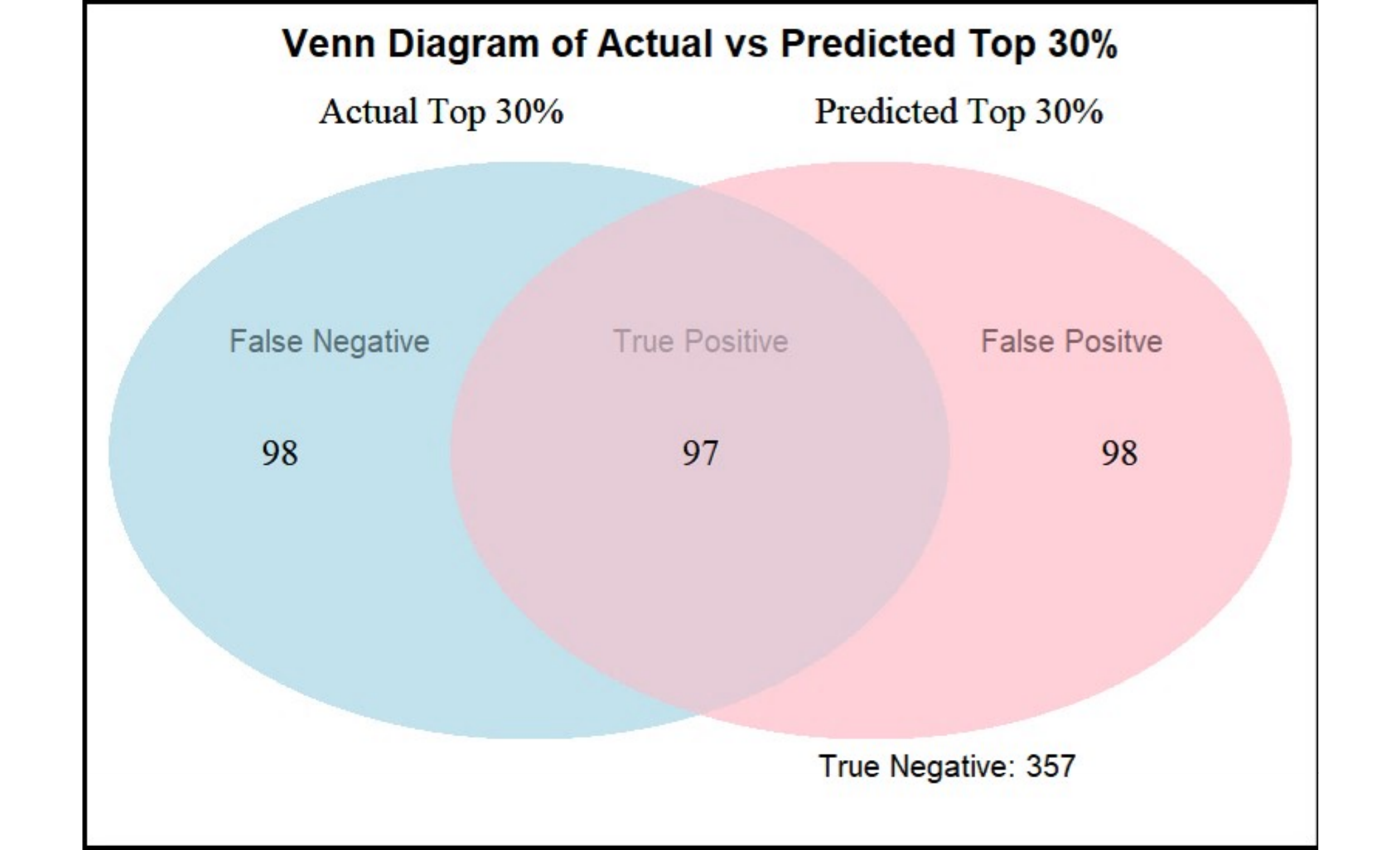}
        \caption{30\% selection threshold}
        \label{fig:venn diagram_3}
    \end{minipage}  
\setcounter{figure}{10}
\setcounter{subfigure}{-1}
    \caption{Venn Diagrams showing actual versus estimated highest yielding lines using a \textbf{(A)} 10\%, \textbf{(B)} 20\%, and \textbf{(C)} 30\% selection threshold using TSC as an estimate for yield ranking.}
    \label{fig:venn diagram}
\end{subfigure}

\begin{subfigure}
\setcounter{figure}{11}
\setcounter{subfigure}{0}
    \centering
    \begin{minipage}[b]{0.5\textwidth}
        \includegraphics[width=\linewidth]{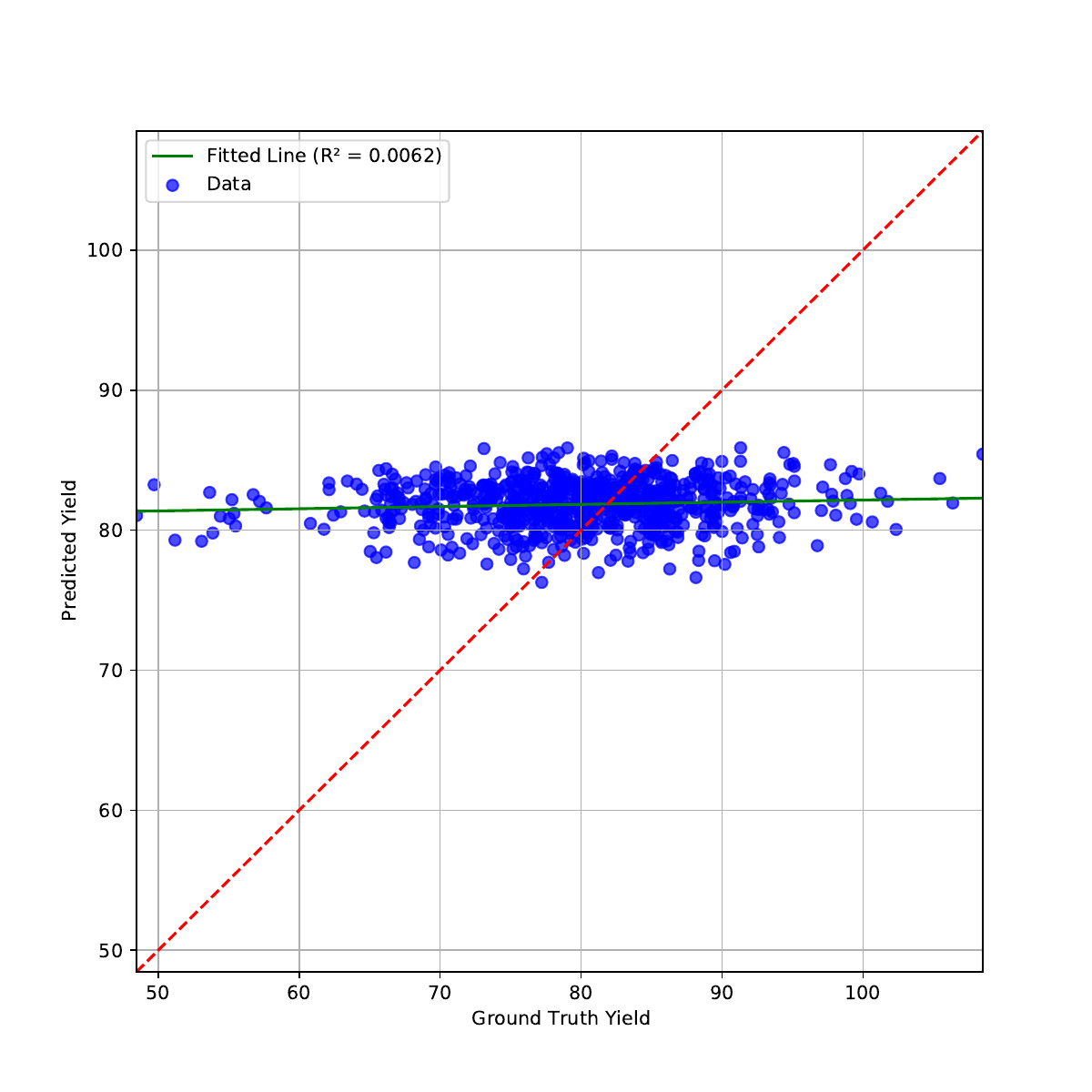}
        \caption{Ground truth yield vs estimated yield using whole dataset.}
        \label{fig:yield_result_1}
    \end{minipage}  
\setcounter{figure}{11}
\setcounter{subfigure}{1}
    \centering
    \begin{minipage}[b]{0.5\textwidth}
        \includegraphics[width=\linewidth]{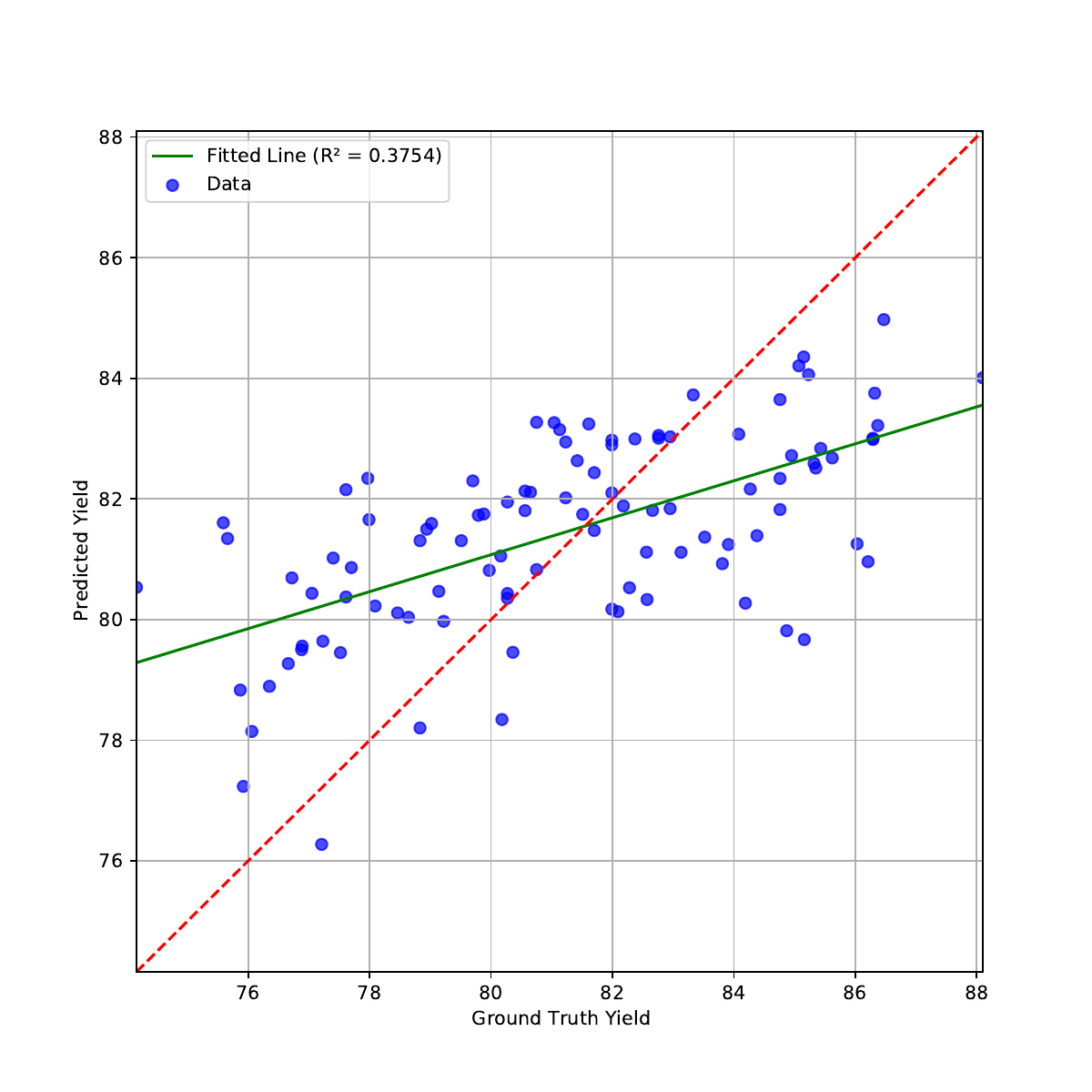}
        \caption{Ground truth yield vs estimated yield using dataset of plots under optimal breeding and imaging situations.}
        \label{fig:yield_result_2}
    \end{minipage}
\setcounter{figure}{11}
\setcounter{subfigure}{-1}
    \caption{Yield estimation results on \textbf{(A)} all and \textbf{(B)} 100 selected plots from the 2023 dataset, which demonstrate the potential effectiveness of our P2PNet-Yield architecture in yield estimation under optimal conditions.}
    \label{fig:yield_result}
\end{subfigure}



\end{document}


\onecolumn
\firstpage{1}

\title[Supplementary Material]{{\helveticaitalic{Supplementary Material}}}

\maketitle

\section{Supplementary Tables and Figures}

\subsection{Improvement of Seed Counting Model's Performance on Corrected Fisheye Images}
\label{sec:sc_improve}

To illustrate the enhancement in seed counting performance with corrected fisheye images, we present seed detection results using two different sets of weights: the original weights and those obtained from training the model on our improved datasets. Figure~\ref{fig:sc_comp_1} shows the results with the original weights, where the model struggles to detect seeds in the corrected fisheye images. In contrast, Figure~\ref{fig:sc_comp_2} demonstrates that the model with the fine-tuned weights successfully detects most seeds in the foreground. This comparison highlights the significant improvement achieved through fine-tuning the model on the enhanced dataset.

\begin{subfigure}
\setcounter{figure}{12}
\setcounter{subfigure}{0}
    \centering
    \begin{minipage}[b]{0.35\textwidth}
        \includegraphics[width=\linewidth]{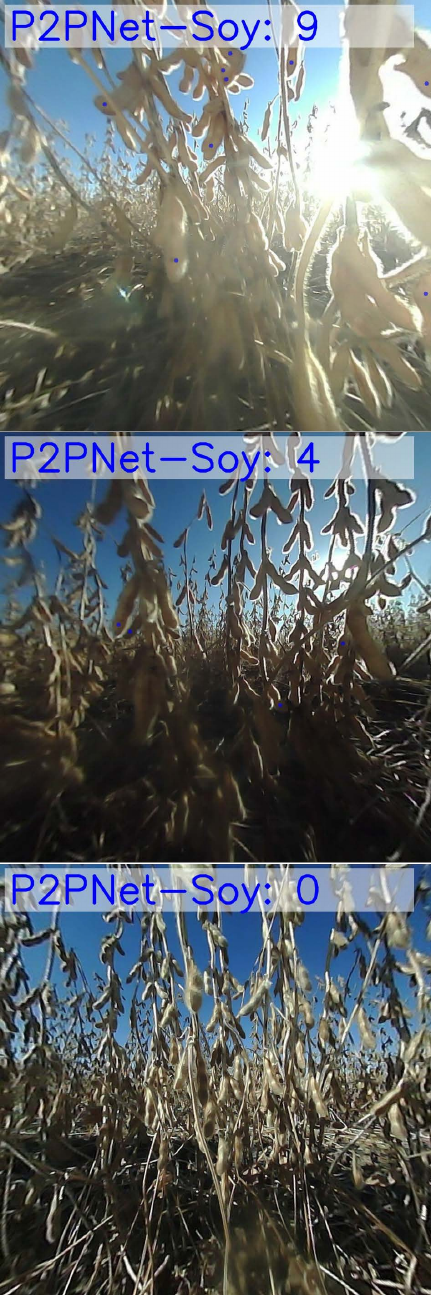}
        \caption{Using original weights}
        \label{fig:sc_comp_1}
    \end{minipage}  
\hspace{0.02\textwidth}
\setcounter{figure}{12}
\setcounter{subfigure}{1}
    \begin{minipage}[b]{0.35\textwidth}
        \includegraphics[width=\linewidth]{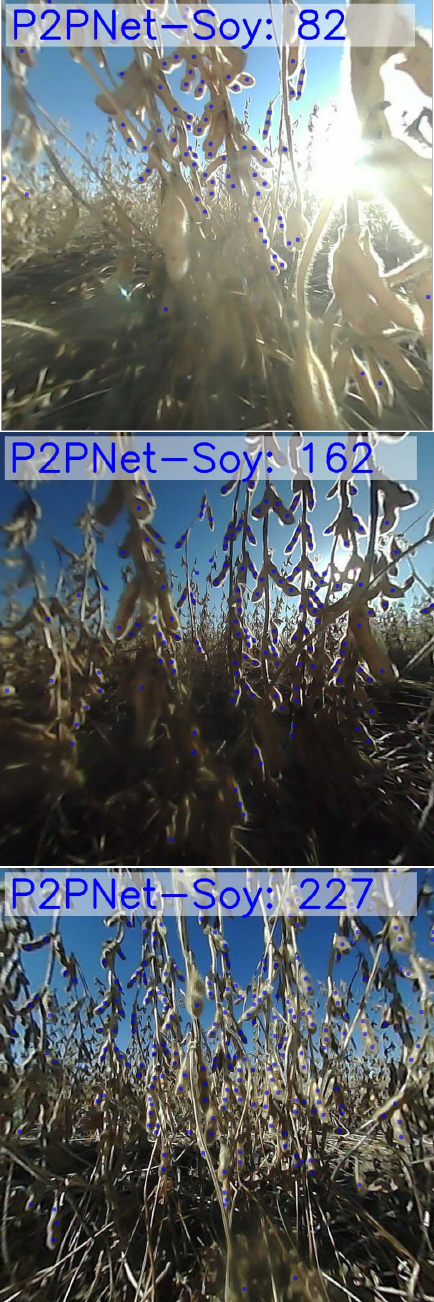}
        \caption{Using our weights.}
        \label{fig:sc_comp_2}
    \end{minipage}

\setcounter{figure}{12}
\setcounter{subfigure}{-1}
    \caption{Comparison of seed detection results using original weights and our weights. \textbf{(A)} Using original weights; \textbf{(B)} Using our weights.}
    \label{fig:sc_comp}
\end{subfigure}